\pdfoutput=1

\documentclass[11pt]{article}

\usepackage[final]{acl}

\usepackage{times}
\usepackage{latexsym}

\usepackage[T1]{fontenc}

\usepackage[utf8]{inputenc}

\usepackage{microtype}

\usepackage{inconsolata}

\usepackage{tabularx}

\usepackage{url}
\usepackage{graphicx}
\usepackage{amsmath}
\usepackage{multirow}
\usepackage{booktabs}
\usepackage{listings}
\usepackage{wasysym}
\usepackage{amssymb}
\usepackage{authblk}

\usepackage{graphicx}

\title{ToolPlanner: A Tool Augmented LLM for Multi Granularity Instructions with Path Planning and Feedback}

\author{\bf Qinzhuo Wu, Wei Liu\thanks{\,\,\,\,Corresponding author.}, Jian Luan, Bin Wang \\
  XiaoMi AI Lab \\
  \texttt{\{wuqinzhuo, liuwei40, luanjian, wangbin11\}@xiaomi.com} \\}

\begin{document}
\maketitle
\begin{abstract}
Recently, tool-augmented LLMs have gained increasing attention.
Given an instruction, tool-augmented LLMs can interact with various external tools in multiple rounds and provide a final answer.
However, previous LLMs were trained on overly detailed instructions, which included API names or parameters, while real users would not explicitly mention these API details.
This leads to a gap between trained LLMs and real-world scenarios.
In addition, most works ignore whether the interaction process follows the instruction.
To address these issues, we constructed a training dataset called MGToolBench, which contains statement and category-level instructions to better reflect real-world scenarios.
In addition, we propose ToolPlanner, a two-stage reinforcement learning framework that utilizes path planning and two feedback mechanisms to enhance the LLM's task completion and instruction-following capabilities. 
Experimental results show that ToolPlanner significantly improves the Match Rate, Pass Rate and Win Rate by \textbf{26.8\%, 20.2\%, and 5.6\%} compared to the SOTA model. 
Human evaluation verifies that the multi-granularity instructions can better align with users' usage habits.
Our data and code are available at \href{https://github.com/XiaoMi/toolplanner}{https://github.com/XiaoMi/toolplanner}.
\end{abstract}

\begin{figure}[!ht]
  \centering
  \includegraphics[width=1\columnwidth]{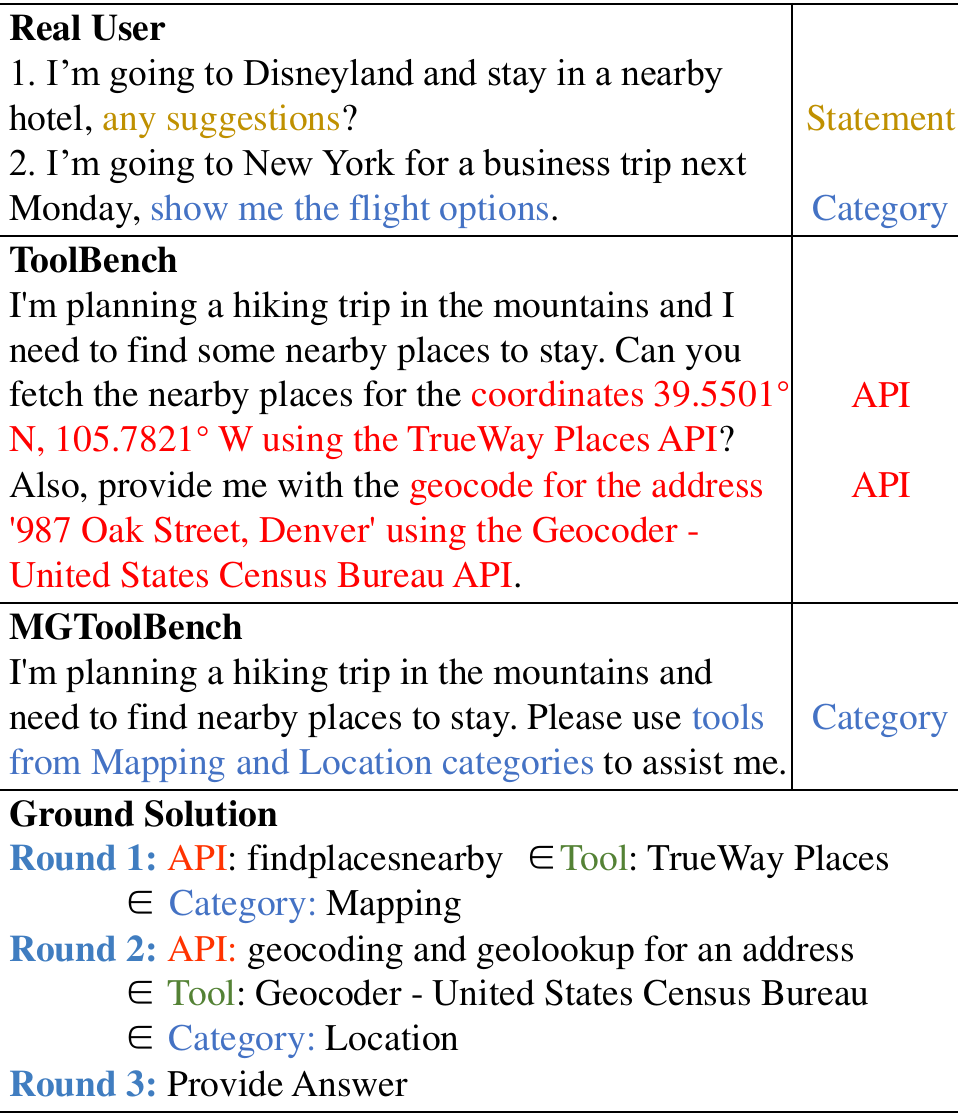}
  \caption{
  Several instructions and their granularity levels from real users, ToolBench, and MGToolBench. Real users tend to provide instructions at a higher level, such as Statement or Category, while ToolBench often consists of more detailed instructions at the API level.
  }\label{figure1-3}
\end{figure}

\section{Introduction}

Recently, tool-augmented large language models (LLMs) have shown their remarkable ability in utilizing various external tools, such as HuggingFace models \cite{shen2023hugginggpt,wu2023visual}, real-world applications \cite{liu2023agentbench,wang2023describe}, and massive APIs \cite{tang2023toolalpaca,liang2023taskmatrix}.
To simulate complex real-world tasks, previous studies have continuously increased the size of the external tool pool and the complexity of task instructions \cite{yang2023gpt4tools,ruan2023tptu,kong2023tptu}. 
LLMs need to break down complex instructions into subtasks, interact with the tools in multiple rounds based on each subtask's requirement, and finally provide a reasonable answer.
However, these instructions often tend to be overly detailed and specific, which differ from real-world scenarios.

After observing online user cases, we noticed that their proposed tasks are similar to the real user examples in Figure \ref{figure1-3}. Users tend to describe their current situation or the category of information they need, rarely mentioning the tools they require, let alone the API names.
Intuitively, users do not care which specific APIs LLM uses to complete their tasks and are unlikely to remember the functions of massive APIs. For example, a ToolBench case like "coordinates 39.5501° N, 105.7821° W" is unlikely to occur in a real-world scenario.

Moreover, previous works focused on whether LLMs could ultimately generate a reasonable answer, while ignoring their ability to follow instructions \cite{wang2023mint}. 
For the ToolBench example in Figure \ref{figure1-3}, the instruction explicitly requires the LLM to complete the task with the "True Way Places" and "Geocoder" tools.
In round 2, if the LLM decides to interact with the "Weather" tool instead of the "Geocoder" tool, it may still generate a valid answer. However, this interaction process does not follow the given instruction, which may result in a decrease in the quality of the final answer.

To address these issues, we constructed a training dataset called MGToolBench using ToolBench as the seed data.
MGToolBench adopts a multi-granularity user instruction mechanism to match user behavior in real-world scenarios.
In addition, we propose ToolPlanner, a two-stage reinforcement learning (RL) framework. In Stage 1, a supervised fine-tuning (SFT) model is used to generate a solution tree for each instruction. 
In Stage 2, two metrics, task completion, and instruction following, are used to score the generated solutions and pair positive and negative responses. We further reinforce the SFT model with pairwise responses as feedback to enhance the model's tool usage ability.
Furthermore, a solution path planning mechanism is used to guide ToolPlanner during its multi-round reasoning process.
The main contributions of this paper can be summarized as follows:

$\bullet$ We constructed a multi-granularity instruction dataset called MGToolBench to reflect real-world scenarios. As far as we know, this is the first study exploring the ability of tool-augmented LLMs to follow instructions of different granularities.

$\bullet$ We proposed ToolPlanner, a two-stage RL framework that utilizes task completion and instruction-following feedback to enhance the model's tool usage abilities. ToolPlanner includes a solution path planning mechanism that provides high-level guidance for the reasoning process.

$\bullet$ Experimental results show that ToolPlanner outperforms existing state-of-the-art models. 
Human evaluation confirms the multi-granularity instruction mechanism's ability to generate instructions that align with real-world scenarios.



\begin{figure*}[!ht]
  \centering
  \includegraphics[width=1\textwidth]{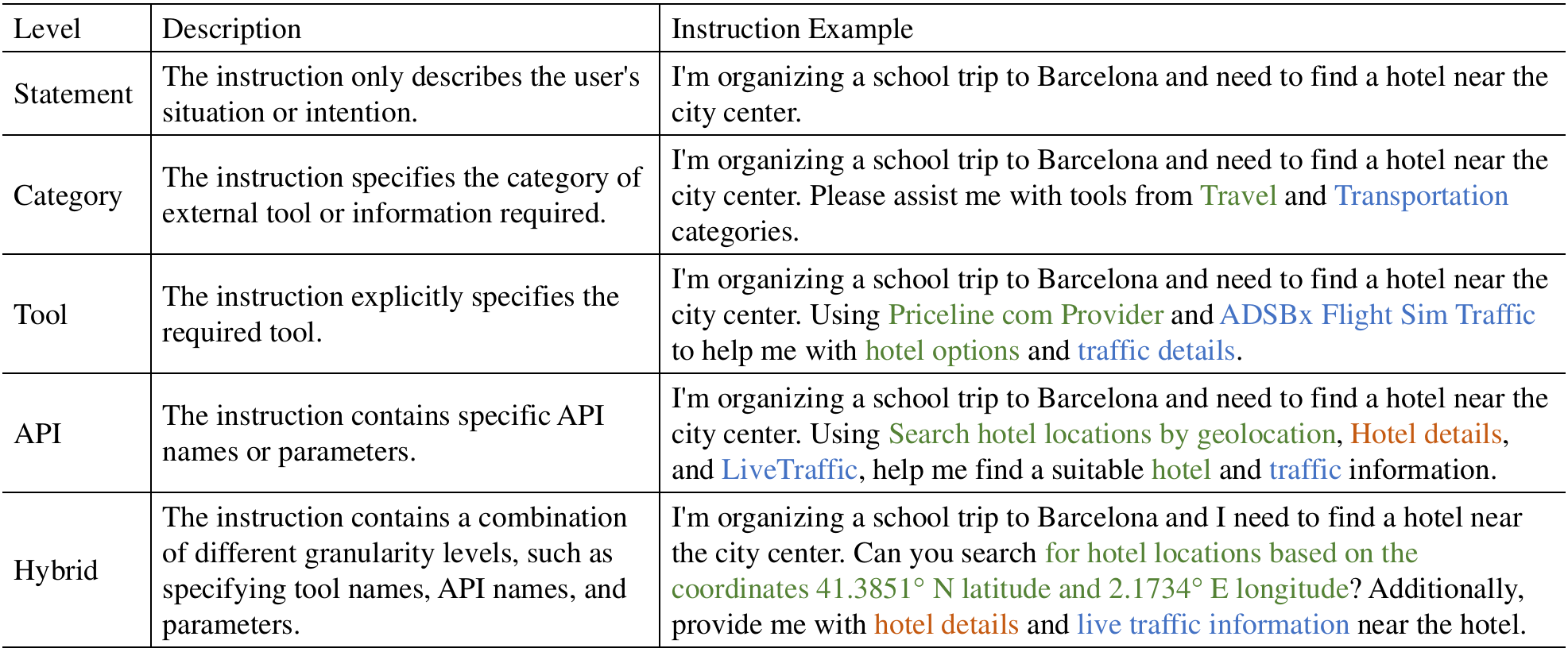}
  \caption{Descriptions and examples of instructions at different granularity levels. 
  }\label{figure1-5}
\end{figure*}

\section{Related Work}
\textbf{Tool-augmented LLMs Datasets:} 
The research community collects diverse datasets to facilitate research on tool-enhanced LLMs.
API-Bank \cite{li2023api} provides a benchmark that includes 264 annotated dialogues and 568 APIs.
APIBench \cite{patil2023gorilla} collects 1,716 machine learning APIs from 3 public model hubs.
ToolBench \cite{qin2023toolllm} provides a high-quality dataset containing 16,464 real-world APIs collected from RapidAPI Hub.  
ToolAlpaca \cite{tang2023toolalpaca} uses real-world APIs similar to ToolBench, with 3,938 instances.
When generating instructions, these works rely heavily on pre-selected tools or APIs. 
This makes the instructions too detailed and inconsistent with the usage habits of real users.

\textbf{Tool-augmented LLMs Framework:} 
Many studies have combined LLMs with massive external tools and APIs to access a wider range of resources and services \cite{parisi2022talm, xu2023tool, liang2023taskmatrix}.
Toolformer \cite{schick2024toolformer} trains LLM to directly generate responses containing API calls in the text.
React \cite{yao2022react} interacts with external tools multiple rounds that follow the "Thought-Action-Action Input-Observation" format.
ToolLLM \cite{qin2023toolllm} uses a tree-based method that can restart the reasoning process from a previous round.
\citet{ye2023large} designed an Elo-based Self-Judgment Mechanism \cite{elo1978rating} that use ChatGPT as a decision maker. 
ToolPlanner uses a two-stage RL framework with solution path planning and two feedback mechanisms to guide the model in its reasoning process.

\textbf{Reinforcement Learning on LLM:} Recently, RL-based fine-tuning mechanisms have been employed to enhance the LLM's generation quality \cite{liu2023rltf, shen2023pangu}.
\citet{yuan2023rrhf} proposed an RRHF paradigm that encourages the model to generate results with better human preferences.
\citet{qiao2023making} enhances the model through feedback derived from tool execution.
We leverage task completion and instruction-following feedback to score and sample pairwise responses at each solution round.

\section{Dataset Construction}
\subsection{Multi-Granularity Instruction Mechanism}\label{section2.1}

To match user behavior in real-world scenarios, we propose a multi-granularity user instruction mechanism. 
We have chosen three intermediate granularity levels that mirror the level of real-world APIs: category, tool, and API. The detailed instructions from seed datasets were set at the hybrid level. The coarse-grained instructions without explicit constraints were set to statement level. As a result, the instructions are now divided into five granularities. 
Descriptions and examples of these instructions are shown in Figure \ref{figure1-5}.

From the figure we can see, the coarser the granularity, the closer the instructions are to the usage habits of real users. Real users are unlikely to provide complex API names and detailed parameters like "41.3851$^{\circ}$ N latitude". 
However, using coarse-grained instructions also results in a larger solution space. For example, in MGToolBench, there are 3 hotel-related tools that contain a total of 60 APIs. For a category-level instruction that requires hotel information, interacting with any of the 60 APIs is considered a reasonable solution. Having only statement or category-level instructions as training data can limit the diversity and usability of the model, as it may only interact with a few common APIs. Therefore, we will use all 5 levels of instructions jointly to build our dataset.

\subsection{MGToolBench Dataset}\label{section2.2}

We use the intra-category multi-tool subset (G3 split) of ToolBench as the seed dataset because it is the largest public tool usage dataset and requires the combined use of multiple tools from different categories to complete complex tasks.
Each seed task contains an instruction and a solution tree, where the rightmost path of the tree is the seed solution.
We removed seed tasks with invalid solution trees, leaving 4,435 remaining tasks. 

\begin{figure}[!tbp]
  \centering
  \includegraphics[width=1\columnwidth]{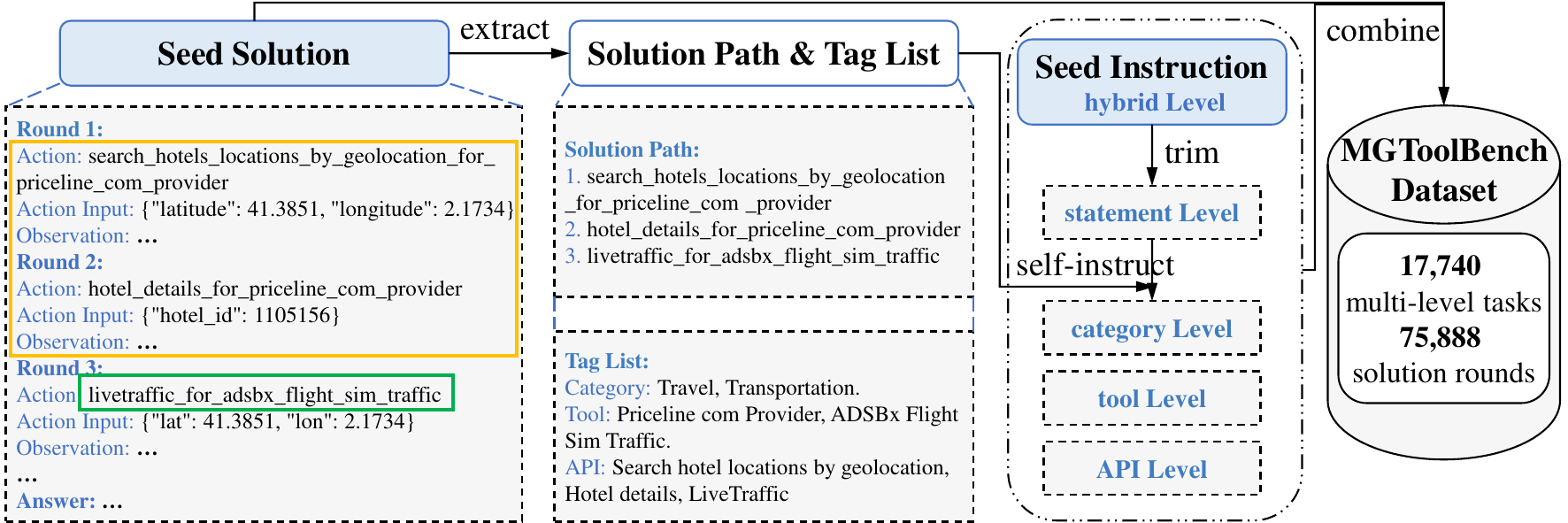}
  \caption{MGToolBench Dataset Pipeline.
  }\label{figure1-6}
\end{figure}

Figure \ref{figure1-6} shows the process of building MGToolBench.
First, for each seed solution, the sequence of interactions with different APIs is extracted to form a solution path. All APIs in the path and their corresponding tools and categories are combined into different levels of tag lists. Then, the seed instructions (i.e., hybrid-level) are trimmed into statement-level instructions by keeping only statements that describe the user's situation. Following Self-Instruct \cite{wang2022self}, statement-level instructions and tag lists at different levels are provided to the GPT-4 model to generate new instructions for the remaining three levels.
After combining these instructions with the seed solution, we had 17,740 multi-level tasks and 75,888 solution rounds, which were used to train the Stage 1 SFT model. Each task consists of an instruction, a tag list, a solution path, and a multi-round solution.\footnote{We only use 50\% of statement-level and category-level tasks as training data for data balance. More details of MGToolBench are shown in Appendix \ref{appendix_3}.}

\begin{figure*}[!htb]
  \centering
  \includegraphics[width=1\textwidth]{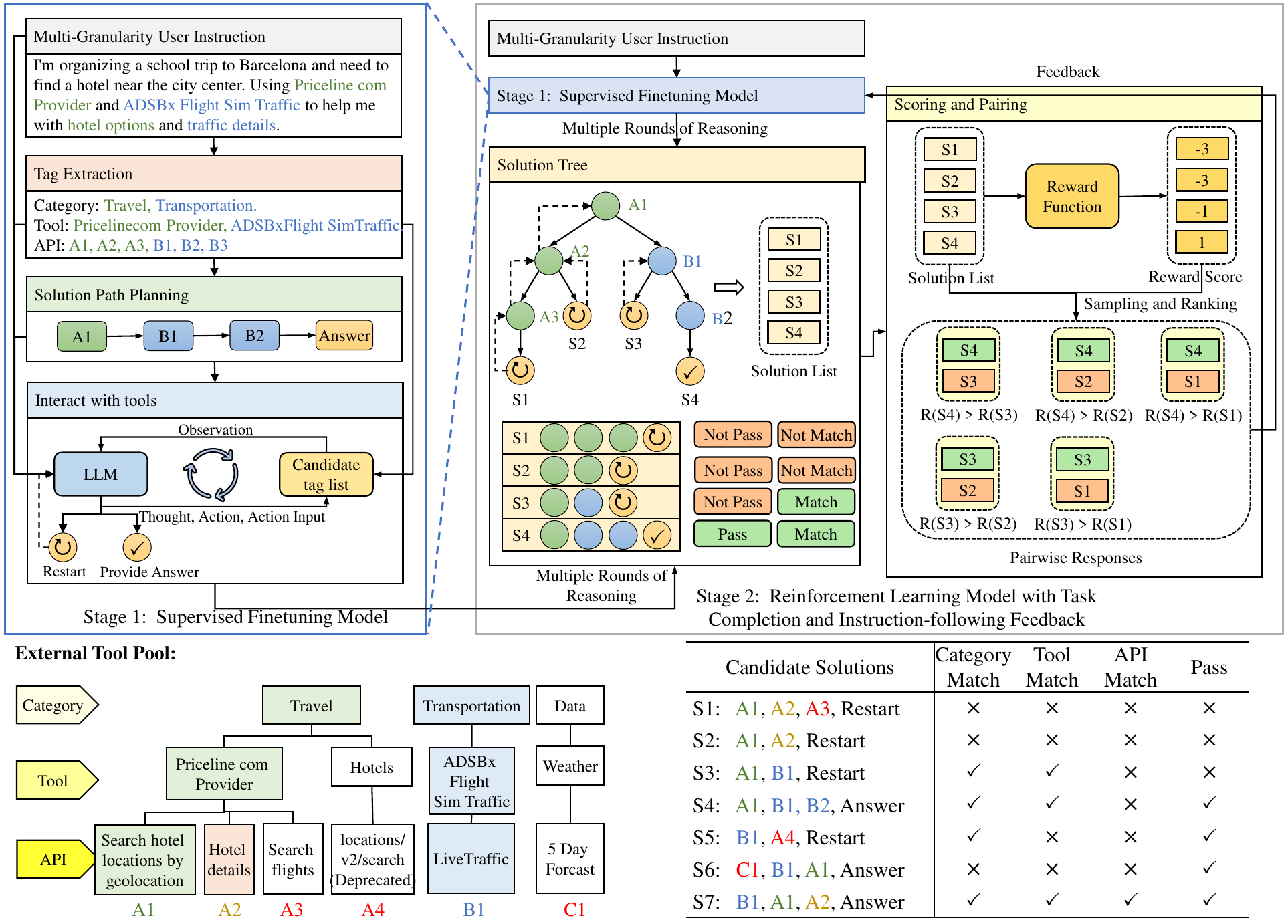}
  \caption{(Top) The overview of our proposed ToolPlanner.
  (Bottom Left): An external tool pool with 6 candidate APIs. (Bottom Right): Results of 7 candidate solutions on our metrics. }\label{figure5-1} %
\end{figure*}

\section{Models}\label{section3}




\subsection{Problem Definition}

The tool-using task can be expressed as a multi-round reasoning process that generates a solution S and a final answer Y based on the given user instruction X. 
As shown in Figure \ref{figure5-1}, ToolPlanner is composed of the Stage 1 SFT model and Stage 2 RL model.
In Stage 1, the SFT model is fine-tuned in a sequence-to-sequence manner, which includes three modules: tag extraction, solution path planning, solution tree generation. In Stage 2, following RRHF \cite{yuan2023rrhf}, we sample pairwise responses with the reward function and use them to continue optimizing the SFT model.\footnote{The prompt for each module is provided in Appendix \ref{appendix_4_1}. The extraction process is provided in Appendix \ref{appendix_RL}. }


\subsection{Stage 1: Supervised Finetuning}\label{section3.1}

\subsubsection{Tag Extraction} 
Given a user instruction, ToolPlanner needs to extract the user's intent and generate a candidate tag list of three granularities. In Figure \ref{figure5-1}, from a tool-level instruction X, we can extract its tool-level list as "Priceline, ADSBx" and its category-level list as "Travel, Transporation". 
At API level, ToolPlanner needs to generate several APIs belonging to these two tools, e.g., "Hotel details"(A2) from "Priceline" or "LiveTraffic"(B1) from "ADSBx".



\subsubsection{Solution Path Planning} 
In this module, given an instruction and a candidate tag list, ToolPlanner generates a complete solution path as a high-level guide for the following process.
As shown in Figure \ref{figure5-1}, 
ToolPlanner believes that it needs to call A1 first, followed by B1, and then B2 to finally generate the answer.

\subsubsection{Solution Tree Generation} 


With the user instruction, the candidate tag list, and the solution path as input, ToolPlanner needs to go through multiple rounds of interaction with external tools to obtain a solution tree. 
Each tool interation round is an intermediate node in the solution tree, including thought, generating an API request, and obtaining an observation. 
The leaf node in the solution tree is a Finish node of the current branch. 
Once LLM generates a Finish node with an answer, the rightmost path of the solution tree is considered the final solution.\footnote{Detailed generation process and a step-by-step inference case are shown in Appendix \ref{appendix_2} and Appendix \ref{appendix_4_3}.}

In Figure \ref{figure5-1}, after interacting with A1,A2,A3,
ToolPlanner decides to restart from the second round. These four rounds form the first solution, S1. After restarting twice more, ToolPlanner ends the tree generation with "Provide Answer". S4 ="A1, B1, B2,$\checked$" is the final solution of the solution tree.

\begin{figure}[!t]
  \centering
  \includegraphics[width=1\columnwidth]{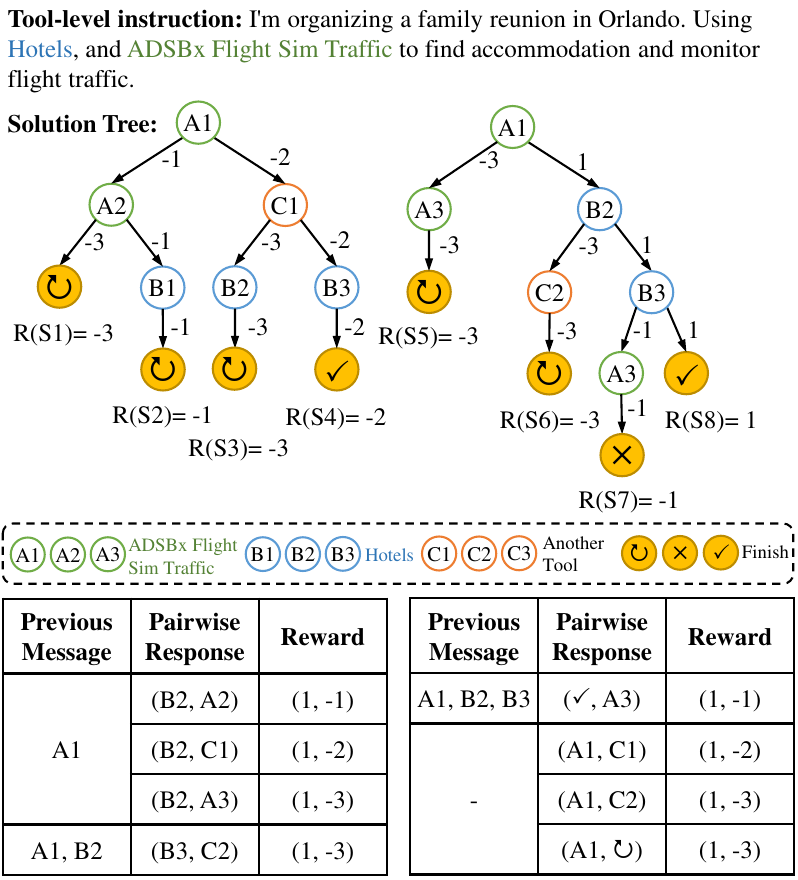}
  \caption{Two solution tree and their pairwise responses for a tool-level instruction. }\label{figure4-1} 
\end{figure}

\subsection{Stage 2: Reinforcement Learning}\label{section3.2}


\subsubsection{Task Completion and Instruction Following Metrics}\label{twometric}
To better evaluate LLM's ability to complete tasks and follow instructions, we propose two metrics:

\textbf{Task completion} measures whether the solution can successfully complete the task.
If the solution finally provides a meaningful answer, mark it as "Pass".
If the solution exceeds the max number of rounds or decides to restart, mark it as "Not Pass". 

\textbf{Instruction-following} measures whether the solution follows the user instruction.
If the solution accesses and only accesses all categories, tools or APIs mentioned in the instruction, it should be marked as "Match" at the corresponding level.

%
%

Figure \ref{figure5-1} shows the evaluation results for 7 candidate solutions.
with different reasoning processes. S4, S5, S6, and S7 end the solution with "Provide Answer" and should be considered "PASS". 
The API-level and hybrid-level instructions provided in Figure \ref{figure1-5} explicitly mention that they should interact with APIs A1, A2, and B1. Therefore, S7 accesses the correct APIs and is considered "Match" at the category, tool, and API level.
B2 accessed by S4 belongs to the correct tool "ADSBx". So S4 is considered "Match" at the tool level and "Not Match" at the API level. Similarly, S5 accesses A4, so S5 only "Match" at the category level.

The goal of these two metrics is to evaluate solutions that not only complete the task but also follow the given instructions. Here, an API or hybrid-level instruction should "Match" on all 3 levels, while a category-level instruction only needs to "Match" at the category level.

Based on the above two metrics, we score the candidate solutions. 
The reward score R(S) of solution S is defined as:
\begin{equation}
\begin{split}
\text{R(S)}\!=\!\begin{cases}
 1 & \text{if S} \in \text{Pass \&  Match} \\
 -1 & \text{if S} \in \text{Not Pass \&  Match} \\
 -2 & \text{if S} \in \text{Pass \&  Not Match} \\
 -3 & \text{if S} \in \text{Not Pass \& Not Match}
\end{cases}
\end{split}
\end{equation}
Take the solution S4="A1,B2,B3,$\checked$" as an example.
For the API-level instruction, R(S4)=-2 because it did not interact with A2. For the tool-level instruction, R(S4)=1 because it "Match" the instruction at the tool level.




\subsubsection{Pairing pairwise responses} \label{sampling}
Intuitively, we expect ToolPlanner to generate solutions with positive rewards. Therefore, we collected pairwise responses to further reinforce the SFT model. 
Specifically, we pair a negative example for each round of positive solutions. 
Two solution rounds can form a pairwise response if they share the same history rounds and their reward is positive and negative, respectively.
Here, for the i-th round $s_i$ of solution S, its reward R($s_i|s_{<i}$) equals the highest reward score among all the solutions to which it belongs. 

In Figure \ref{figure4-1}, S8="A1,B2,B3,$\checked$" is a positive solution with 4 rounds. 
In round 3, B3 and C2 share a common history "A1,B2" and R(B3|A1,B2)=1 is greater than R(C2|A1,B2)=-3. We consider (B3,C2|A1,B2) as a pairwise response.
In round 1, there is no negative round for node A1. We will sample a round and ensure it belongs to a negative solution. E.g., the pairwise responses could be (A1,C1|-) and (A1,$\circlearrowright$|-). Finally, for the t-th round $s_t$, we have at least a response pair $(s_t^1,s_t^2|s_{<t})$, where R($s_t^1|s_{<t}$) > R($s_t^2|s_{<t}$).

\subsection{Training} 
In Stage 1, 
we use cross-entropy loss to train the SFT model to generate the candidate tag list C, solution path P, solution tree S, and answer Y.
\begin{equation}
{\mathcal{L}}= \sum_{t} \log \mathbf{P}(s_t, {\rm Y}| s_{<t}, \rm P, C, X).
\end{equation}
In Stage 2, we use pairwise responses to further finetune the solution tree generation module in the SFT model. In t-th round, for a pairwise response $(s_t^1,s_t^2|s_{<t})$, the ranking loss is defined as:
\begin{equation}\
{\mathcal{L}}_{rank}=\!\!\!\! \sum_{{\rm R}(s_t^1)> {\rm R}(s_t^2)}\!\! {\rm min} (0,{\rm P}(s_t^2)-{\rm P}(s_t^1)).
\end{equation}
The final loss function $\mathcal{L}$ is a combination of the cross-entropy loss and ranking loss:
\begin{equation}
\setlength\belowdisplayskip{5pt}
\begin{split}
{\mathcal{L}}_{l}&=- \sum_{t}\log \mathbf{P}(s_t|s_{< t},\rm X)),\\
\mathcal{L}&={\mathcal{L}}_{l}+\beta{\mathcal{L}}_{rank}.\\
\end{split}
\end{equation}
Here, $\beta$ is a hyperparameter.



\begin{table*}[!tb]
\centering
\resizebox{\textwidth}{!}{
\begin{tabular}{l|c|c|cc|ccc|ccc|c|cccccc|cccccc}
\hline
\hline

  \multirow{3}{*}{Model}& & \multicolumn{10}{c|}{Match Rate (\%)}                                             & \multicolumn{6}{c|}{Pass Rate (\%)} & \multicolumn{6}{c}{Win  Rate (\%)} \\ \cline{2-24}
                        & Instruction & Cate   & \multicolumn{2}{c|}{Tag} & \multicolumn{3}{c|}{API}  & \multicolumn{3}{c|}{Hybrid}  & Avg.  & State & Cate     & Tag     & API     & Hybrid    & Avg.  & State & Cate     & Tag     & API     & Hybrid   & Avg.  \\   \cline{2-12} \cline{18-18}  \cline{24-24} 
                       & Tag Level  & C  & C         & T         & C     & T     & A     & C     & T     & A   &  &  &    &     &       &       &       &       &       &       &   & &   \\ \hline
ChatGPT-Chain & FewShot    & 4        & 22            & 17       & 13        & 11    & 7   & 21         & 15     & 8  &13.1 & 42 & 25       & 36   & 31  & 20  &30.8
 & 50  & 28       & 44   & 34 & 32  &37.6   \\
ChatGPT-Tree & FewShot      & 3        & 20            & 14       & 15        & 14    & 8   & 16         & 11     & 4  &11.7 & 55  & 32       & 46   & 39  & 38 &42.0 & 57   & 43       & 48   & 46 & 45     &47.8 \\
GPT4-Chain & FewShot      & 8        & 51            & 42       & 25        & 23    & 18  & 39         & 34     & 21 &29.0 & 71 & 60       & 68   & 62  & 41  &60.4  & \textbf{82} & \textbf{80}       & 71   & 74 & 54   &72.2   \\
GPT4-Tree & FewShot & 10       & 44            & 38       & 23        & 21    & 14  & 45         & 40     & 24 &28.8 & 67 & 63       & 68   & 64  & 45  &61.4 & 74   & 79       & 66   & 76 & 63    &71.6  \\ \hline
ToolLLaMA-Chain & SFT   & 5        & 48            & 35       & 30        & 24    & 13  & 41         & 34     & 9  &26.6 & 61 & 30       & 41   & 20  & 14  &33.2
 & 69   & 41       & 43   & 28 & 47    &45.6 \\
ToolLLaMA-Tree & SFT    & 8        & 33            & 22       & 26        & 19    & 9   & 38         & 28     & 13 &21.8 & 68  & 60       & 74   & 66  & 53 &64.2   & 73  & 64       & 77   & 73 & 69 & 71.2 \\  
ToolLLaMA-2-7b & SFT & 10 & 47 & 41 & 33 & 29 & 16 & 50 & 45 & 26 & 33 & 82 & 74 & \textbf{85} & 74 & 60 & 75 & 79 & 76 & 77 & 73 & 69	& 74.8\\
\hline
\textbf{ToolPlanner} & SFT      & \textbf{59}       & \textbf{61}            & \textbf{57}       & \textbf{64}        & \textbf{61}    & \textbf{52}  & \textbf{60}         & \textbf{52}     & \textbf{37} &\textbf{55.8} & \textbf{88} & \textbf{89}       & 84   & \textbf{83}  & \textbf{78}   & \textbf{84.4}   & 78  & 79       & \textbf{77}   & \textbf{80}    & \textbf{75} & \textbf{77.8}\\  \hline \hline
\end{tabular}
}
    \caption{Match Rate, Pass Rate and Win Rate of baselines on user instructions at different granularity levels. 
    }
    \label{result1}
\end{table*}

\begin{table}[!htb]
\centering
\resizebox{\columnwidth}{!}{
\begin{tabular}{l|c|c|c}
\hline \hline
Model                  & Method        & Pass        & Win         \\ \hline
GPT-4 Turbo Chain \cite{guo2024stabletoolbench} & FewShot       & 52.5        & 67.2        \\
GPT-4 Turbo Tree \cite{guo2024stabletoolbench} & FewShot       & 66.1        & 60.7        \\
ToolLLaMA v2 Chain \cite{guo2024stabletoolbench} & SFT           & 33.9        & 24.6        \\
ToolLLaMA v2 Tree \cite{guo2024stabletoolbench} & SFT           & 53.6        & 50.8        \\
AnyTool \cite{du2024anytool}            &  Agent   & 63.2        & -           \\
Sum2Act \cite{liu2024summary}            &  Agent & 74          & 50.6        \\ \hline
\textbf{ToolPlanner}            & SFT           & \textbf{78} & 75          \\ \hline \hline
\end{tabular}
}
    \caption{Pass Rate and Win Rate of previous studies. 
    }
    \label{result1_1}
\end{table}

\section{Experiment}
\subsection{Dataset}
We use the G3 split of ToolBench \cite{qin2023toolllm} as a seed to construct the MGToolBench dataset, which contains 75,888 solution steps for Stage 1 SFT model training.
To obtain more negative solutions, we regenerate the solution tree for the multi-level instructions using the SFT model.
Finally, we have 98,950 paired responses for stage 2 RL model training.
We use the official G3 split test set with 100 hybrid-level tasks for better comparison. 
Therefore, there was no overlap between the training set and the test set. 
We use the multi-granularity instruction mechanism to generate test instructions at the other four levels. See more details  Appendix \ref{appendix_data}.

\subsection{Settings}\label{metric}
\textbf{Baselines.}
We compare our proposed ToolPlanner with the following baselines:
ChatGPT (gpt-3.5-turbo-16k) \cite{chatgpt2022} is one of the most advanced LLMs currently available.
GPT4 (gpt-4-0314) \cite{openai2023gpt4} is a more powerful and intelligent LLM with stronger tool usage capabilities. 
ToolLLaMA is a tool-use framework based on LLaMA-7B \cite{touvron2023llama}, which includes a separate API retriever and has been fine-tuned with the ToolBench dataset. 
%

\noindent
\textbf{Decoding Methods.}
1.Chain-based Method: Following ReAct \cite{yao2022react}, CoT@N independently runs chain-based reasoning N times until it finds a solution path that passes the task.
2.Tree-based Method: Following DFSDT \cite{qin2023toolllm}, LLM treats ReAct's multi-step reasoning (Thought-Action-Observation) as a round and performs depth-first reasoning process in a tree structure.

\noindent
\textbf{Main Metric.}
1.Match Rate calculates the proportion of solutions that successfully match user instructions at a certain tag level.
2.Pass Rate\cite{qin2023toolllm} calculates the proportion of solutions that successfully complete the task with a reasonable answer.
3.Win Rate uses ToolEval \cite{qin2023toolllm} to calculate the ratio at which ChatGPT prefers the generated answers over the golden answers.

\noindent
\textbf{Human Evaluation Metric.}
1.Plausibility measures whether an instruction is fluent, complete, and makes sense in describing a user's intent.
2.Conciseness measures whether an instruction is concise. 
3.Relevance measures whether an instruction's instruct clause is clear and relevant to its statement. 
4.Realness measures whether an instruction aligns with the real-world scenarios.



\noindent
\textbf{Implementation Details.}
To ensure fair comparisons, we maintain consistent hyperparameters across all the baselines and our models. ToolPlanner chose LLaMA-7B as the backbone model just like ToolLLaMA. 
In Stage 1, models are trained for 3 epochs on 75,888 instruction-solution rounds. 
In Stage 2, the model was trained for 2 epochs on 98,950 pairwise responses. See more metric details in Appendices \ref{appendix_metric1} and \ref{appendix_human} and more hyperparameter details in Appendix \ref{appendix_1}.



\subsection{Results and Discussions}
The main experimental results for Match Rate, Pass Rate, and Win Rate are presented in Table \ref{result1} and Table \ref{result1_1}. From the table we can observe that:


$\bullet$ Models with a tree-based decoding method perform better on Pass and Win Rate, but worse on Match Rate because tree-based method sacrifices instruction-following ability to complete the task. 

$\bullet$ ToolPlanner achieves a significantly higher Match Rate than other baselines, which we attribute to our multi-granularity instruction mechanism and instruction-following feedback.
ToolPlanner explicitly considers whether the tool meets the instruction requirements in each interaction round, leading to a strong instruction-following ability.

$\bullet$ ToolPlanner significantly outperforms all the other baselines in Pass Rate and Win Rate. 
This can be attributed to our solution path planning mechanism and task completion feedback, which encourage the model to generate a complete solution with a reasonable answer.

$\bullet$ Statement-level instructions do not have tag requirements, so evaluating their Match Rate is unnecessary. ToolPlanner exceeds GPT-4 by 17\% in Pass Rate and is comparable to GPT-4 in Win Rate. This proves that ToolPlanner has a higher task completion rate and delivers high-quality answers for statement-level instructions, which closely resemble real-world scenarios.

$\bullet$ Overall, ToolPlanner improves the Match Rate, Pass Rate, and Win Rate by \textbf{+26.8\%}, \textbf{+20.2\%} and \textbf{+5.6\%} compared to ToolLLaMA-Tree.
This indicates that ToolPlanner can flexibly complete instructions of different levels, and provide high-quality answers for instructions that are close to real-world scenarios.

\begin{table}[!tb]
  \centering
  \resizebox{\columnwidth}{!}
  {
    \begin{tabular}{l|cccc}
    \hline
    Level & Plausibility  & Conciseness & Relevance & Realness  \\
    \hline
    Statement & \textbf{2.85} & \textbf{2.99} & - & \textbf{2.94}  \\
    Category    & 2.73 & 2.97 & \textbf{2.60} & 2.89 \\
    Tool   & 2.58 & 2.98 & 2.50 & 2.68 \\
    API     & 2.01 & 2.72 & 2.13 & 2.19 \\
    Hybrid    & 2.62 & 2.01 & 2.46 & 2.60 \\
    \hline
    \end{tabular}
  }
  \caption{Human evaluation results. These metrics are rated on a 1-3 scale (3 for the best). }\label{table_human1} 
\end{table}

\begin{table*}[!tb]
\centering
\resizebox{1\textwidth}{!}{
\begin{tabular}{ll|ccccc|ccccc}
 \hline \hline
\multirow{3}{*}{Model}           & \multirow{3}{*}{Method}                                   & \multicolumn{5}{c|}{Human}  & \multicolumn{5}{c}{Human Statement}   \\ \cline{3-12}
                &                                          & \multicolumn{3}{c}{Match}     & Pass & Win & \multicolumn{3}{c}{Match}     & Pass & Win  \\  \cline{3-5} \cline{8-10}
                &                                          & Category & Tool & API &      &     & Category        & Tool & API &      &      \\  \hline
GPT4            & FewShot                                  & 45       & 39   & 9   & 70   & 78  & 42              & 36   & 8   & 55   & 76   \\
ToolLLaMA       & ToolBench SFT                  & 46       & 36   & 8   & 69   & 75  & 40              & 34   & 10  & 56   & 73   \\
ToolLLaMA-v2-7b & ToolBench SFT                    & 59       & 54   & 13  & 74   & 79  & 57              & 48   & 11  & 71   & 75   \\
ToolLLaMA-SFT   & MGToolBench SFT w/o RL & 25       & 25   & 11  & 29   & 31  & 31              & 27   & 9   & 36   & 30   \\
ToolLLaMA-2-SFT & MGToolBench SFT w/o RL & 49       & 46   & 15  & 70   & 64  & 40              & 34   & 10  & 66   & 63   \\
ToolPlanner     & MGToolBench SFT                  & 60       & 54   & 20  & 75   & 77  & 55              & 50   & 17  & 84   & 75  \\
 \hline \hline
\end{tabular}}
  \caption{Model performance on 100 human-written instructions. }\label{table_human3} 
\end{table*}

\begin{table*}[!tb]
\centering
\resizebox{\textwidth}{!}{
\begin{tabular}{l|c|c|cc|ccc|ccc|c|cccccc|cccccc}
 \hline \hline
\multirow{3}{*}{\qquad Model \quad} &    & \multicolumn{10}{c|}{Match Rate (\%)}                                             & \multicolumn{6}{c|}{Pass Rate (\%)} & \multicolumn{6}{c}{Win Rate (\%)} \\ \cline{2-24}
                        & Instruction Level & Cate   & \multicolumn{2}{c|}{Tag} & \multicolumn{3}{c|}{API}  & \multicolumn{3}{c|}{Hybrid}  & Avg.  & State & Cate     & Tag     & API     & Hybrid    & Avg.  & State & Cate     & Tag     & API     & Hybrid   & Avg.  \\   \cline{2-12} \cline{18-18}  \cline{24-24} 
                       & Tag Level  & C  & C         & T         & C     & T     & A     & C     & T     & A   &  &  &    &     &       &       &       &       &       &       &   & &   \\ \hline
\multicolumn{2}{l|}{ToolLLaMA-Tree  }          & 8  & 33        & 22        & 26    & 19    & 9     & 38    & 28    & 13  & 21.8 & 68  & 60    & 74    & 66    & 53  & 64.2 & 73 & 64    & \textbf{77}    & 73    & 69  &71.2\\ 
\multicolumn{2}{l|}{ToolLlama-Tree-Finetune} &37	&49	&45	&56	&50	&35	&53	&49	&32 & 45.1 &63 & 52 &46	&32	&29	&44.4   &70 &48	&46	&38	&33 &47.0
\\ \hline
\multicolumn{2}{l|}{ToolPlanner w/o both RL feedbacks  }      & 30 & 53        & 50        & 50    & 48    & 37    & 48    & 43    & 31 & 43.3  & 56   & 50    & 43    & 33    & 31 &42.6  & 59   & 49    & 44    & 35    & 37   &44.8\\ 
\multicolumn{2}{l|}{ToolPlanner w/o  task completion	}& 37 & 45    & 40    & 26    & 24    & 21    & 40    & 35    & 22  &32.2 & 46   & 50    & 38    & 20    & 20 & 34.8 & 44   & 49    & 38    & 22    & 24 &35.4\\
\multicolumn{2}{l|}{ToolPlanner w/o instruction-following  }  & 16    & 15    & 14    & 10    & 8    & 7    & 14    & 12    & 7 & 11.4 & 84   & 87    & 81    & 68    & 76 &79.2 & 80   & \textbf{ 79}    & 74    & 63    & 68 &72.8\\ \hline
\multicolumn{2}{l|}{ToolPlanner w/o Tag \& Path} & 5  & 35        & 25        & 27    & 22    & 14    & 32    & 24    & 11  & 21.7 & 88   & 71    & 84    & 72    & 66  &76.2 & 81   & 67    & 76    & 69    & 64   &71.4\\
\multicolumn{2}{l|}{ToolPlanner w/o Tag  }     & 27 & 32        & 28        & 37    & 30    & 14    & 36    & 30    & 15 & 27.7 & 69   & 73    & 74    & 67    & 71  &70.8 & 63   & 68    & 67    & 58    & 66   &64.4\\
\multicolumn{2}{l|}{ToolPlanner w/o Path }   & 34 & 32        & 22        & 41    & 37    & 26    & 36    & 27    & 19  & 30.4 & 82   & 83    & \textbf{86}    & 76  & 77  &80.8 & 73    & 73    & \textbf{77}    & 76    & 70   &73.8\\ \hline
\multicolumn{2}{l|}{ToolPlanner }              & \textbf{59} & \textbf{61}        & \textbf{57}        & \textbf{64}    & \textbf{61}    & \textbf{52}    & \textbf{60}    & \textbf{52}    & \textbf{37} & \textbf{55.9} & 88   & \textbf{89}    & 84    & \textbf{83}   & \textbf{78}  &\textbf{84.4}  & 78  & \textbf{ 79}    & \textbf{77}    & \textbf{80}    & \textbf{75}  &\textbf{77.8}
\\ \hline \hline
\end{tabular}
}
    \caption{Ablation study on reducing tag extraction, solution path planning mechanisms and two RL feedbacks.}
    \label{result3}
\end{table*}

\subsection{Human Evaluation}

\textbf{Quality of multi-granularity instructions:} 
To evaluate our multi-granularity instruction mechanism, we conducted a human evaluation. We randomly selected 100 instructions with different levels and asked three annotators to rate them on a 1-3 scale across four metrics.
As shown in  Table \ref{table_human1}:

$\bullet$ Fine-grained instructions, such as hybrid-level and tool-level, achieve lower plausibility and realness. This is because their descriptions are too detailed and lengthy, and may contain API and tool names with oddly formatted.

$\bullet$ Statement-level and category-level instructions achieve better performance in each metric. This is because they are very brief, fluent, and close to real-world scenarios. However, our dataset contains only 36 categories, and only using these instructions as training data may lack diversity.

$\bullet$ Overall, this human evaluation confirms that the coarse instructions generated by the multi-granularity instruction mechanism are not only fluent and relevant to the task, but also more aligned with real world scenarios. They are an important supplement to the original hybrid-level instructions.
We recommend using a combination of all 5 levels of instructions. Appendix \ref{appendix_human2} provide an additional human evaluation of the generated answers.

\textbf{Performance on human-written instructions:} 
To verify the effectiveness of the model on real-world user instructions, we ask human annotators to handwrite new instructions based on the tag lists from the test set as a real-world test set. We believe that these human-written instructions can better reflect real-world scenarios.
Specifically,  there are two types of human-written instructions: 

1. Human: annotators write instructions based only on the tag lists.

2. Human-Statement: annotators can use both statement sentences and tag lists.

We use different fine-tuning baselines to generate solutions based on these human-written instructions. The evaluation results are shown in Table \ref{table_human3}. These results show that our ToolPlanner still outperforms the rest baselines on Match Rate and Pass Rate, and is slightly worse than GPT-4 on Win Rate. We believe this proves the effectiveness of ToolPlanner in real-world user instructions.

\subsection{Ablation Study}
Table \ref{result3} shows the results of several ablation experiments.
We can observe that:

\noindent\textbf{Effect of tag extraction and solution path planning mechanisms:} 
Without solution path planning, the Match Rate of "ToolPlanner w/o Path" would decrease by 25.5\%. Without tag extraction, "ToolPlanner w/o Tag"  decreases by 28.2\%, 13.6\%, and 13.4\% in its three metrics. Removing both of these mechanisms further reduces the model’s performance, but it still outperforms ToolLLaMA-Tree, which once again proves the effectiveness of using reinforcement learning.
These results demonstrate the effectiveness of solution path planning, which guides the model to think and reason globally, and make wiser decisions. Additionally, removing the tag extraction causes a bigger decrease because the tag extraction provides a multi-level candidate tag list that helps the model select tools. 

\noindent\textbf{Effect of two RL feedbacks:} Without instruction-following feedback, the Match Rate of ToolPlanner would decrease by about 44.5\%. Without task completion feedback, the performance of ToolPlanner would decrease on all three metrics. 
These results confirm that two RL feedbacks can improve the model's ability to follow instructions and generate final answers.
After using Stage 1 data to further fine-tune the ToolLlama-7B. The performance of "ToolLlama-Tree-Finetune" is similar to "ToolPlanner w/o both RL feedbacks". 


\begin{table}[!tb]
\centering
\resizebox{1\columnwidth}{!}{
\begin{tabular}{l|ccc}
 \hline \hline
\multirow{2}{*}{Model} &\multicolumn{3}{c}{Hybrid} \\ \cline{2-4}
            & Percentage      & Recall      & F1     \\ \hline
Retriever@1 &  \textbf{0.86}   & 0.305  & 0.4503 \\
Retriever@3 &  0.6734 & 0.7092 & 0.6908 \\
Retriever@5 & 0.4949 & 0.8617 & 0.6287 \\ \hline
\textbf{Tag Extraction}         &  0.8486  & \textbf{0.8546}  & \textbf{0.8516} 
\\ \hline \hline
\end{tabular}
}
\caption{The performance of Tag Extraction and Retriever in generating candidate lists.  The full version of multi-level instructions is shown in Table \ref{result7}. 
}

    \label{result5}
\end{table}

\begin{figure}[t]
  \centering
  \includegraphics[width=1\columnwidth]{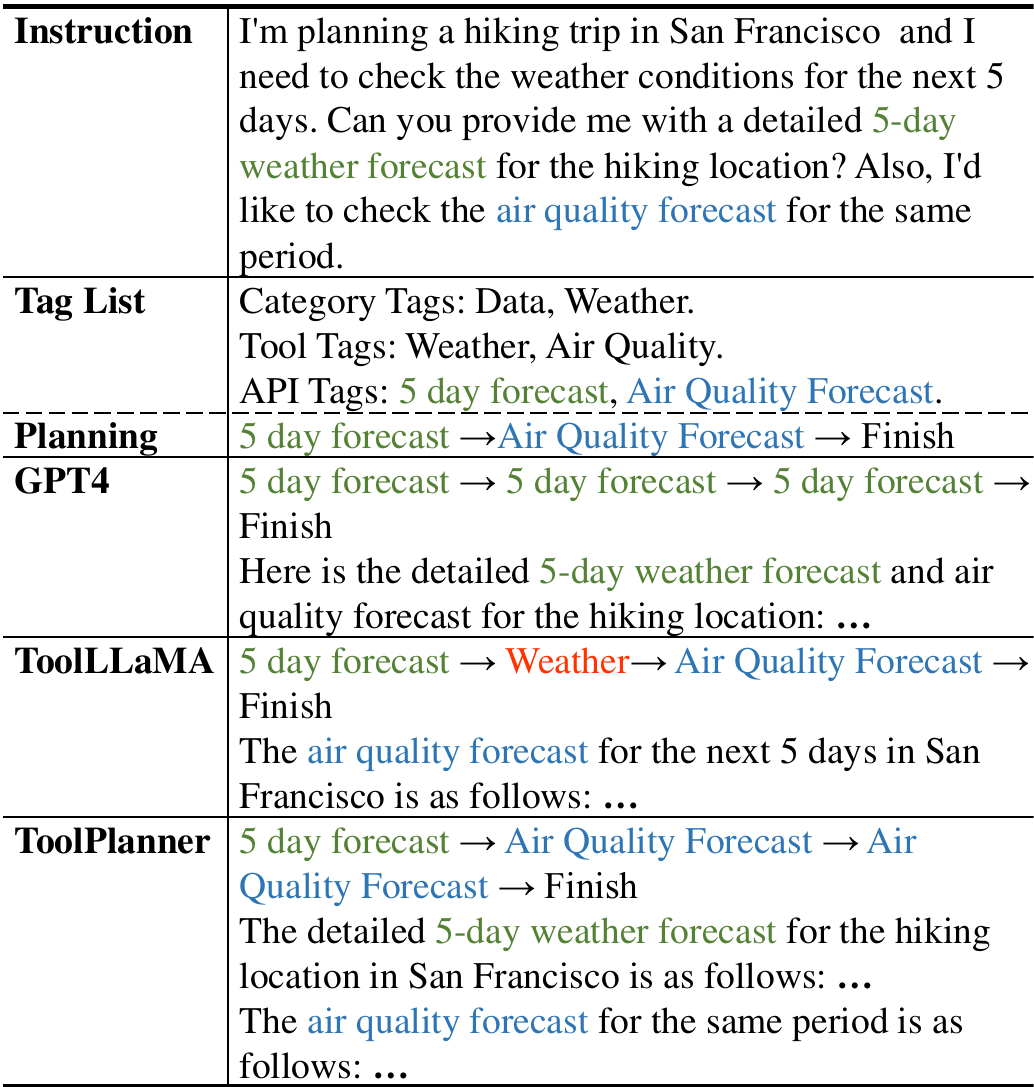}
  \caption{A case of generated solutions by GPT-4-Tree, ToolLlama-Tree and ToolPlanner. }\label{figure6-1}

\end{figure}

\noindent\textbf{Comparison between Tag Extraction and Retriever:} 
Previous works use a dense retriever, following Sentence BERT \cite{reimers2019sentence}, to select the top-K related APIs.
We compare it with our tag extraction mechanism and set K to $(1,3,5)$.
As shown in Table \ref{result5}, the tag extraction mechanism consistently outperforms the dense retriever in different levels of user instructions.
As K increases, the retriever's precision decreases while its recall increases. This is because the number of tools and APIs involved in user instructions is uncertain. In contrast, the tag extraction mechanism can better adapt to different user instructions and provide user intent for the following steps.



\subsection{Case Study}

Figure \ref{figure6-1} shows a case generated by GPT-4, ToolLlama and our ToolPlanner.
After two incorrect API requests,
GPT-4 successfully called the "5 day forecast" API.
However, it ignored the task of air quality prediction. 
ToolLlama mistakenly called the "Weather" API from the "Ambee Air Quality" tool, as the term "Weather" and the task instruction are semantically related. This ultimately caused the model to forget to provide weather forecasts in the answer.
With the tag extraction and solution path prediction mechanism, our ToolPlanner can now predict the tools required to complete the entire task on a global scale, rather than just selecting the tools most relevant to the instruction. 
By using the reinforcement learning, ToolPlanner can learn to avoid using tools that are not mentioned in the instructions, and it can encourage the reasoning process to ultimately provide an answer.

\section{Conclusion}
In this work, we propose ToolPlanner, a two-stage RL framework that utilizes task completion feedback and instruction-following feedback to enhance LLMs' reasoning and tool usage abilities. 
Additionally, we constructed a training dataset called MGToolBench, which uses multi-granularity instructions to simulate the usage habits of real users.
Experimental results show that ToolPlanner significantly improves the Match Rate, Pass Rate and Win Rate by \textbf{26.8\%, 20.2\%, and 5.6\%}. 
Human evaluation verifies that the multi-granularity instruction mechanism can generate instructions that better align with user habits.

By addressing the challenges of tool-augmented LLMs in following user instructions at different granularities, our framework shows strong instruction-following and task-completion abilities. 
We hope that MGToolBench can serve as a helpful resource for simulating real-world scenarios and help future research improve the practical application ability of tool-augmented LLMs.


\section*{Limitations}
ToolPlanner's reasoning process takes too many rounds. Due to the use of a tree-like inference structure, each instruction may require 4-30 rounds to generate a solution tree (as shown in Appendix \ref{appendix_1}), and each round requires 3 interactions (Thought, Action, Action Input) with the LLM. 
This limitations will be the focus of our future work.

\section*{Ethics Statement}
This paper was conducted in accordance with the ACM Code of Ethics. The ToolBench dataset used in this work is publicly available \cite{qin2023toolllm}, and our MGToolBench dataset is constructed using publicly available platforms and data sources, ensuring that there are no privacy issues or violations. All data used in our research was obtained following legal and ethical standards, and we do not collect any personally identifiable information. 

In the human evaluation, we hired 3 crowd workers from the crowdsourcing platform without any discrimination. For the instruction human evaluation, we provided them with 5 instructions of different granularity in MGToolBench. For the answer generation human evaluation, we provided them with the corresponding instructions and the final answers generated by different baselines. We paid these workers no less than RMB 100 per hour.

\newpage
\bibliography{acl_latex}

\newpage
\appendix
\section{HyperParameter Settings}\label{appendix_1}
We present the hyperparameters for Stage 1 SFT model and Stage 2 RL model in Table \ref{parameters}. We choose LLaMA-7B as the backbone model, just like ToolLLaMA, to ensure a fair comparison. The learning rate is first warmed up to the set value, and then linearly decayed to 0. We use 8 80GB Nvidia A100 GPUs for fine-tuning, typically costing 8 hours for Stage 1 and 30 hours for Stage 2. 

\begin{figure}[!htb]
  \centering
  \includegraphics[width=1\columnwidth]{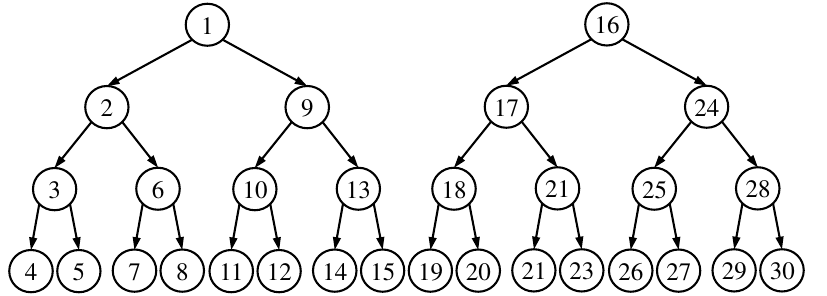}
  \caption{During the reasoning process, ToolPlanner generates the solution tree in a depth-first manner.}\label{figure4-4}

\end{figure}

Following \cite{chen2023extending}, we set the maximum sequence length to 8192. This is because the prompt used to generate the solution tree will exceed 4,500 characters after adding user instructions and descriptions of Tools and APIs, as described in Appendix \ref{appendix_4_3}. $\beta$ in our loss function is 1 \cite{yuan2023rrhf,liu2023rltf}.  The maximum number of steps for each solution is 12. Since each round contains 3 steps (Thought, Action, Action Input), the maximum number of steps for each solution is 4. For the chain-based method, N of CoT@N  is set to 5. For the tree-based method, each node in the solution tree has at most 2 children. At most two solution trees are generated in each reasoning process. Therefore, in a reasoning process, if the model generates two full solution trees, the maximum number of reasoning rounds is 30, as shown in Figure \ref{figure4-4}.

\begin{table}[!htb]
    \centering
\resizebox{\columnwidth}{!}{
    \begin{tabular}{|l|c|c|}
        \hline
Hyperparameter & Stage 1  & Stage 2\\
&SFT model& RL model \\
        \hline
epoch & 2 & 3 \\
batch\_size & 2 & 1 \\
learning\_rate & 5e-5 & 2e-5\\
warmup\_ratio & 0.04 & 0.03\\
weight\_decay & 0.0 & 0.0 \\
optimizer & Adam & Adam \\
max\_sequence\_length & 8192 & 8192 \\
GPUs  & 8 & 8 \\
         \hline
    \end{tabular}}
    \caption{Hyperparameters.}
    \label{parameters}
\end{table}

\section{Details for MGToolBench Dataset} \label{appendix_3}
\subsection{Data Statistics}\label{appendix_data}
We report the statistics of seed data in Table \ref{dataset1},
which is the intra-category multi-tool instruction subset from ToolBench \cite{qin2023toolllm}. It is the most challenging subset in ToolBench. It requires the combined use of multiple tools from different categories, which helps to reflect complex real-world scenarios. We removed seed tasks that did not provide a proper candidate tag list or had an invalid solution tree, leaving 4,435 remaining tasks. 

We use the official test set for the intra-category multi-tool instruction subset from ToolBench. The test set consists of 100 tasks. We build test instructions using the same approach as building multi-granularity instructions in the training dataset.
Specifically, we set the original test instructions at the hybrid level, and used these instructions, their corresponding tag lists, and instruction generation prompts to feed into the GPT-4 model to generate test instructions at the other three levels.

\begin{table}[!htb]
    \centering
\resizebox{\columnwidth}{!}{
    \begin{tabular}{lccccc}
        \hline
        &\multicolumn{2}{c}{Dataset }  &  \multicolumn{3}{c}{Tag Vocab }\\ 
        \cmidrule(r){2-3} \cmidrule(r){4-6}
 & Train & Test & Category & Tool & API\\
        \hline
Size & 4,435 & 100 & 36 & 240 & 1,332\\
         \hline
    \end{tabular}}
    \caption{The statistics of seed dataset.}
    \label{dataset1}
\end{table}

We report the statistics of our constructed dataset, MGToolBench, in Table \ref{table_1_2}.
 \begin{table}[!htb]
    \centering
    \resizebox{\columnwidth}{!}{
\begin{tabular}{l|c|c|c}
\hline
  \multirow{2}{*}{Number}           & \multirow{2}{*}{ToolBench} & Multi-Level & \multirow{2}{*}{MGToolBench} \\ 
            & & Instruction &\\ \hline
Task        & 4,435   &  4,435 & 21,981       \\ \hline
Instruction & 4,435   & 17,740  & 87,924       \\ \hline
Step        & 18,972  &  75,888 & 331,340      \\ \hline
\!Pairwise Response\! & -   & - & 98,950      \\ \hline
\end{tabular}}
\caption{Statistical information of the MGToolBench dataset.}
\label{table_1_2}
\end{table}

\begin{table*}[!htb]
    \centering
    \resizebox{\textwidth}{!}{
    \begin{tabular}{|l|}
        \hline
"query": "I'm planning a trivia night and I need a variety of questions. Can you provide me with a music \\ trivia question from the Music Trivia API? Also, fetch me a random trivia question from the Trivia by \\ API-Ninjas API. Thanks!",\\
"query\_id": 101,\\
"relevant APIs": [ [ "Music Trivia", "/getgamelevel"],
\ [ "Trivia by API-Ninjas", "/v1/trivia"] ],\\
"api\_list": [\\
\quad \{"category\_name": "Media", "tool\_name": "Music Trivia", "api\_name": "/getgamelevel" \},\\
\quad \{"category\_name": "Gaming", "tool\_name": "Trivia by API-Ninjas", "api\_name": "/v1/trivia" \},\\
\quad \{"category\_name": "Social", "tool\_name": "Chuck Norris", "api\_name": "/jokes/categories" \},\\
\quad \{"category\_name": "Social", "tool\_name": "Chuck Norris", "api\_name": "/jokes/random" \},\\
\quad \{"category\_name": "Social", "tool\_name": "Chuck Norris", "api\_name": "/jokes/search" \}
]\\
         \hline
    \end{tabular}
    }
    \caption{An instruction example from ToolBench dataset.}
    \label{data_1}
\end{table*}

\begin{table*}[!htb]
    \centering
    \resizebox{\textwidth}{!}{
    \begin{tabular}{l|l}
         \hline
         \hline
role & content \\
\hline
\hline
system & System Prompt ... ... \\
\hline
user  &  I'm hosting a trivia night with a focus on numbers, and I need some interesting facts. Can \\
& you fetch a math fact about the number 1729, a trivia fact about the number 42, and a fact \\
& about the year 1492? It would be great if you could also provide a trivia question from the  \\
& Trivia by API-Ninjas API. Begin! \\
\hline
assistant & \{
            "name": "get\_math\_fact\_for\_numbers",
            "arguments": "\{   "number": "1729" \}"
          \}\\
         \hline
function & \{"error": "", "response": "\{'text': 'the smallest natural number representable in two different \\
& ways as a sum of two positive cubes, as Ramanujan stated on the spot', 'number': 1729, \\
& 'found': True, 'type': 'math'\}"\}\\
\hline
assistant & \{
            "name": "get\_math\_fact\_for\_numbers",
            "arguments": "\{   "number": "42" \}"
          \}\\
         \hline
function & \{"error": "", "response": "\{'text': 'the number of spots (or pips, circular patches or pits) on \\
& a pair of standard six-sided dice', 'number': 42, 'found': True, 'type': 'trivia'\}"\} \\
\hline
assistant & \{
            "name": "get\_math\_fact\_for\_numbers",
            "arguments": "\{   "number": "1492" \}"
          \}\\
         \hline
function & \{"error": "", "response": "\{'date': 'October 12', 'text': "Christopher Columbus'  expedition \\
    & makes landfall in the Caribbean and lands on Guanahani, but he believes he has reached \\
& the East Indies", 'number': 1492, 'found': True, 'type': 'year'\}"\} \\
\hline
assistant & \{
            "name": "v1\_trivia\_for\_trivia\_by\_api\_ninjas",
            "arguments": "\{\}"
          \}\\
         \hline
function & \{"error": "", "response": "[\{'category': 'geography', 'question': "Australia:- Australia Produces\\
    &   Over 90\% Of One Of The Earth's Semi Precious Stones Which One ", 'answer': 'Opal'\}]"\} \\
\hline
assistant & \{
            "name": "Finish",
            "arguments": "\{  "return\_type": "give\_answer",   "final\_answer": \\
    &   "Here are some interesting facts: \\
    &  1. The number 1729 is the smallest natural number representable in two different ways as a \\
    &  sum of two positive cubes, as Ramanujan stated on the spot.\\
    &  2. The number 42 is the number of spots on a pair of standard six-sided dice.\\
    &  3. In the year 1492, Christopher Columbus' expedition made landfall in the Caribbean and \\
    &  landed on Guanahani, but he believed he had reached the East Indies.\\
    &   And here's a trivia question:\\
    &  Category: Geography\ Question: Australia produces over 90\% of one of the Earth's semi-\\
    &  precious stones. Which one? Answer: Opal." \} \\

         \hline
         \hline
    \end{tabular}
    }
    \caption{A solution example from ToolBench dataset.}
    \label{data_5}
\end{table*}

To generate multi-granularity instructions, first, we collected tag lists with different levels of granularity from the original instructions. 
Then, we provided the tag lists and statements to the GPT-4 model to generate new instructions at at three other levels.
We obtained 75,888 solution steps with multi-level instructions, and used them to train the Stage 1 SFT model.

To generate more pairwise responses, we first used the Stage 1 SFT model to regenerate solution trees for user instructions at each granularity level. Then, we used the reward function to score these solution trees and sample pairwise responses for each solution step. We only consider steps with a reward score of 1 as positive responses and pair them with negative steps that have the same history.
In total, we collected a training dataset containing 98,950 pairwise responses.

The detailed process of multi-granularity instruction generation is provided in Appendix \ref{instruction_generation}, and the detailed process of pairwise responses extraction is provided in Appendix \ref{appendix_RL}.

\subsection{ToolBench}

We chose ToolBench as our main dataset because it is currently the largest publicly available dataset. However, our multi-granularity instruction mechanism can also be applied to other public datasets, like APIBench\cite{patil2023gorilla} and ToolAlpaca\cite{tang2023toolalpaca}. These datasets may use different types of tags to construct multi-granularity instructions, such as domains, model names, App names, etc. Therefore, our method has broad applicability and scalability.

The instructions in ToolBench consist of three types: single-tool instructions (G1), intra-category multi-tool instructions (G2), and intra-collection multi-tool instructions (G3). As described in Section \ref{section2.2}, we only use the intra-category multi-tool instruction subset from ToolBench as a seed to construct the MGToolBench dataset. In G3 subset, they randomly select 2-5 tools from the same collection and sample at most 3 APIs from each tool to generate the instructions. 

We focus on the G3 subset because it is the most challenging subset in ToolBench. It requires the combined use of multiple tools from different categories, which helps to reflect complex real-world scenarios. In addition, ToolBench provides 5,000 instruction-solution pairs for the G3 subset, which are used to train ToolLlama. These solutions were collected by ChatGPT and employ a reasoning strategy based on Depth First Search-based Decision Tree.

\subsubsection{Instruction Format}

Table \ref{data_1} shows an instruction example that requires obtaining "music trivia question" and "random trivia question". The ground-truth solution is to access the "Music Trivia" and "Trivia by API-Ninjas" tools, as provided by the "relevant APIs". In addition, each instruction also has an "API list" as its external tool pool, which contains several APIs that are most relevant to this instruction. However, we only use "relevant APIs" to build our tag list. This is because it is inappropriate to call the Chuck Norris tool when the instruction explicitly requires the use of the Music Trivia API and the API-Ninjas API.

\subsubsection{Solution Format}
Table \ref{data_5} shows a solution example from ToolBench dataset. In this example, the model interacted three times with the tool "numbers" using different parameters, and once with the tool "trivia\_by\_api\_ninjas". Combining the responses from the four API requests, the model generated the final answer.

\begin{table}[!htb]
    \centering
    \resizebox{\columnwidth}{!}{
    \begin{tabular}{l|l}
        \hline
Category & Sports\\
Tool & Live Sports Odds \\
APIs & /v4/sports/{sport}/odds, \\
& /v4/sports/{sport}/scores, /v4/sports \\
         \hline
Category & Food\\
Tool & Tasty \\
APIs & recipes/auto-complete, tags/list, \\
& recipes/list-similarities, recipes/list, \\
& feeds/list, recipes/detail (Deprecated), \\
& tips/list, recipes/get-more-info \\
         \hline
Category & Social\\
Tool & Chuck Norris \\
APIs & /jokes/random, /jokes/search, \\
& /jokes/categories \\
         \hline
Category & Data\\
Tool & Weather \\
APIs & Current Weather Data of a location., \\
& 5 day Forecast, 16 Day Forecast, \\
& Severe Weather Alerts, 120 Hour Forecast, \\
& 1 Hour / Minutely Forecast (Nowcast) \\
\hline
    \end{tabular}
    }
    \caption{Several cases for Category/Tool/API Format.}
    \label{data_3}
\end{table}

\begin{table}[!htb]
    \centering
    \resizebox{\columnwidth}{!}{
    \begin{tabular}{|l|}
        \hline
My family and I are planning a ski trip to Aspen. Can \\ 
you provide us with the current weather conditions \\ and 
a 120-hour forecast for the coordinates 39.2\textbackslash u00\\ b0N  and 106.8\textbackslash u00b0W? Also, let us know if there \\  are any active weather alerts in the region.  Finally, \\ 
recommend some popular ski resorts and slopes in\\  Aspen. \\
         \hline
I'm planning a company event and I want to create a \\ 
fun and engaging atmosphere. Fetch the latest memes \\  from 
the Programming Memes Reddit API and show \\  me some rising 
popular posts from Reddit. \\ Additionally, check   if a 
specific is username available \\ on all platforms  using the Check Username API. \\
         \hline
I need the exchange rate from EUR to GBP.  \\ Additionally, retrieve a comment from Deezer with \\  the id '5555' and a trivia fact about the year 2022. \\
\hline
    \end{tabular}
    }
    \caption{Several examples of user instructions.}
    \label{data_4}
\end{table}

\begin{table*}[!htb]
    \centering
    
    \resizebox{\textwidth}{!}{
    \begin{tabular}{l|l}
         \hline
         \hline
Hybrid-level Instruction & I need to plan a beach party for my company. Can you give me the 5-day weather \\
& forecast for Miami and suggest some cocktail recipes that complement the \\
& weather?  Also, provide me with the detailed recipe for a cocktail with the ID 45. \\
\hline
\hline
API-level Input  &  I need to plan a beach party for my company. \\  & API: get\_5\_day\_forecast, list\_of\_cocktails, detailed\_cocktail\_recipe\_by\_id. \\
\hline
API-level Instruction & I need to plan a beach party for my company. Using get\_5\_day\_forecast, \\
& list\_of\_cocktails, and detailed\_cocktail\_recipe\_by\_id APIs, provide me with \\
& weather  predictions and cocktail ideas.\\
\hline
\hline
Tool-level Input & I need to plan a beach party for my company.\\
& Tool: weather, the\_cocktail\_db, the\_cocktail\_db.\\
         \hline
Tool-level Instruction & I need to plan a beach party for my company. Using Weather and The\_Cocktail\_DB \\
& to provide weather updates and cocktail ideas.\\
\hline
\hline
Category-level  Input & I need to plan a beach party for my company.\\
& Category: Data, Food, Food. \\
         \hline
Category-level Instruction\!\!& I need to plan a beach party for my company. Please provide me with relevant \\
& information using tools from Data and Food categories. \\
         \hline
         \hline
    \end{tabular}
    }
    \caption{An example seed instruction and its multi-granularity instructions. }
    \label{data_2}
\end{table*}

\begin{table*}[!htb]
    \centering
    \resizebox{\textwidth}{!}{
    \begin{tabular}{|l|}
        \hline
You are a research assistant. Please generate a coarse-grained tool usage instruction based on detailed user \\ instructions for a tool usage task. You should not provide a detailed task description and need to include \\ the api name in the simplified instruction.\\
\\
Example1: \\
System: I'm planning a surprise party for my best friend's birthday.\\
Category: Food, Weather, Sports.\\
Answer: I'm planning a surprise party for my best friend's birthday. Please help me find some information \\ with tools from Food, Weather, Sports categories.\\
\\

Example2: \\
System: I'm organizing a charity event for my company and we need some assistance.\\
Category: Translation, Business.\\
Answer: I'm organizing a charity event for my company and we need some assistance. Using tools from \\ Translation and Business category, and give me some ideas.\\
\\
Now, Please make the simplified answer of below requests.\\

System: \{request\}\\
Answer:\\
         \hline
    \end{tabular}
    }
    \caption{The prompt for category-level instruction generation.}
    \label{prompt_1}
\end{table*}

\begin{table*}[!htb]
    \centering
    \resizebox{\textwidth}{!}{
    \begin{tabular}{|l|}
        \hline
         You are a research assistant. Please generate a coarse-grained tool usage instruction based on detailed user \\ instructions for a tool usage task. You should not provide a detailed task description and need to include \\ the api name in the simplified instruction.\\
\\
Example1: \\
System: I'm planning a surprise party for my best friend's birthday.\\
Tool: The Cocktail DB, Weather, Free NBA.\\
Answer: I'm planning a surprise party for my best friend's birthday. Using The Cocktail DB, Weather and\\  Free NBA to find me some cocktial recipe, weather forecast and basketball information.\\
\\

Example2: \\
System: I'm organizing a charity event for my company and we need some assistance.\\
Tool: Microsoft Translator Text, MyMemory - Translation Memory.\\
Answer: I'm organizing a charity event for my company and we need some assistance. Using these two\\  tools, Microsoft Translator Text, MyMemory - Translation Memory, and give me some ideas.\\
\\
Now, Please make the simplified answer of below requests.\\

System:  \{request\}\\
Answer:\\
         \hline
    \end{tabular}
    }
    \caption{The prompt for tool-level instruction generation.}
    \label{prompt_2}
\end{table*}
\begin{table*}[!htbp]
    \centering
    \resizebox{\textwidth}{!}{
    \begin{tabular}{|l|}
        \hline
         You are a research assistant. Please generate a coarse-grained tool usage instruction based on detailed user \\ instructions for a tool usage task. You should not provide a detailed task description and need to include \\ the api name in the simplified instruction.\\
\\
Example1: \\
System: I'm planning a surprise party for my best friend's birthday.\\
API: Detailed Cocktail Recipe by ID, 16 Day Forecast, Get a Specific Game.\\
Answer: I'm planning a surprise party for my best friend's birthday. Using Detailed Cocktail Recipe by \\ ID, 16 Day Forecast, Get a Specific Game to find me some cocktial recipe, weather forecast and basketball \\ information.\\
\\

Example2: \\
System: I'm organizing a charity event for my company and we need some assistance.\\
API: Languages, search translations.\\
Answer: I'm organizing a charity event for my company and we need some assistance. Using these two \\ APIs,  Languages, search translations, and give me some ideas.\\
\\
Now, Please make the simplified answer of below requests.\\

System: \{request\}\\
Answer:\\
         \hline
    \end{tabular}
    }
    \caption{The prompt for api-level instruction generation.}
    \label{prompt_3}
\end{table*}

\subsection{MGToolBench Dataset}

\subsubsection{Conflict between Instructions and Real Users}

When constructing data, ToolBench will provide several tools, APIs, and their documentation to allow ChatGPT to generate user instructions. Therefore, ChatGPT tends to directly copy the API name or introduction from the documentation, rather than using a more natural description.

The conflict between the generated instructions and user habits comes from two aspects:
\begin{itemize}
    \item 1. many API names are designed for developers and do not conform to the usage habits of real users. 
Table \ref{data_3} shows several tools along with their corresponding APIs and categories. 
Some API names do not conform to the natural language format, some API names may overlap with the names of other tools, and some API functions are difficult to tell from their names.

\item 2. Because the tools are given first and then the instructions are generated, the instructions are described in too much detail, even including specific APIs and parameters used.
\end{itemize}

In real-world scenarios, users' descriptions of tasks are often more ambiguous. They do not provide their latitude and longitude in order to ask about the weather. 

From Table \ref{data_4} we can see, the weather tool has an API called "120 Hour Forecast", which causes the model to use a similar name when generating instructions. Real users are more likely to ask for "weather forecast for the next few days."

\subsubsection{multi-granularity instruction generation}\label{instruction_generation}

In ToolBench, each instruction has a corresponding API list. Because when building data, ToolBench first samples several APIs, and then lets ChatGPT generate instructions that can use these APIs at the same time.
Table \ref{data_1} shows an example of the data instructions in ToolBench, which provides "relevant APIs" for each instruction. We simply use these API names, tool names, and retrieve the category names to which the tools belong as tag lists.

We set the original seed instructions at the hybrid level, and used these instructions, their corresponding tag lists, and instruction generation prompts to feed into the GPT-4 model to generate instructions at the other three levels.

Table \ref{data_2} shows the detailed process of multi-granularity instruction generation. For a seed instruction from ToolBench, the statement is "I need to plan a beach party for my company," and its corresponding external tools are "weather" and "the\_cocktail\_db."

With the statement and the API list (get\_5\_ day\_forecast, list\_of\_cocktails, detailed\_cocktail \_recipe\_by\_id) as input, along with API-level prompts in Table \ref{prompt_1}, we can generate API-level instructions. Tool-level instructions and category-level instructions are generated using a similar method, as shown in Table \ref{data_2}.

In this way, we collected 17,740 multi-granularity instructions based on 4,435 seed instructions.
We only used the intra-category multi-tool instruction subset from ToolBench as a seed to construct the MGToolBench dataset. Similarly, we only used 100 test sets from G3\_instruction to construct our multi-granularity instruction test set. We built test instructions using the same approach as in the training set.

\subsubsection{Prompt Design}
In this section, we show the details of the prompt design for generated multi-granularity instructions based on seed instructions and tag lists. 

The prompts of the category-level, tool-level and API-level instruction generation are shown in Table \ref{prompt_1}, \ref{prompt_2}, \ref{prompt_3}, respectively.

\newpage

\begin{table*}[!htb]
    \centering
    \resizebox{\textwidth}{!}{
    \begin{tabular}{|l|}
        \hline
         You are a helpful assistant and good planner. Your job is to find which APIs assistant can use by given the \\ seed task and tools. \\
First I will give you the a user request and its corresponding tools as the seed task, and your job start.\\
Here are some examples of human request and corresponding tools: \\
\\

System: I'm planning a surprise birthday party for my best friend and I want to create a special cocktail  \\ menu. Can you provide me with a list of cocktail recipes, including their names, images, and detailed \\ recipes? Additionally, fetch some relevant images of cocktails to design personalized party invitations. \\
Tag: Thought: Cate\_Tag: Food, Food, Data.\\
Tool\_Tag: the\_cocktail\_db, the\_cocktail\_db, web\_search.\\
API\_Tag: list\_of\_cocktails, detailed\_cocktail\_recipe\_by\_id, imagesearch.\\
\\

System: I am planning a family vacation to New York and need to book round-trip flights and a rental car. \\ Using search\_round\_trip, search\_results\_request, and livetraffic APIs, help me find suitable flights and a \\ rental car.\\
Tag: Thought: Cate\_Tag: Travel, Travel, Transportation.\\
Tool\_Tag: priceline\_com\_provider, priceline\_com\_provider, adsbx\_flight\_sim\_traffic.\\
API\_Tag: search\_round\_trip, search\_results\_request, livetraffic.\\
\\


Now, Please make the API using plan of below requests and tools.\\
System: \{request\} \\
Tag:\\ 
         \hline
    \end{tabular}
    }
    \caption{The prompt for tag extraction.}
    \label{prompt_4}
\end{table*}

\begin{table*}[!htb]
    \centering
    \resizebox{\textwidth}{!}{
    \begin{tabular}{|l|}
        \hline
         Assume that you play a role of tool using planner, I would give you a user request and its corresponding \\ tag list, and you should help me to plan the tool using solution path.\\
Here are some examples of human request and corresponding tool using solution path: \\
\\

System: I'm planning a fun-filled weekend with my family and I want to start it off with a good laugh. \\ Using socialgrep, programming\_memes\_reddit, find me some entertaining content. \\
Cate\_Tag: Data, Data, Entertainment.\\
Tool\_Tag: socialgrep, socialgrep, programming\_memes\_reddit.\\
API\_Tag: post\_search, comment\_search, get\_all\_memes.\\
Solution\_Path:\\
Thought: get\_all\_memes\_for\_programming\_memes\_reddit, post\_search\_for\_socialgrep, comment\\ \_search\_for\_socialgrep, comment\_search\_for\_socialgrep, Finish.\\
\\

System: Please suggest a fun fact about a random year and a random NBA player's statistics. Using \\ get\_random\_fact, get\_all\_stats, and jokes\_search to find an interesting fact and NBA player statistics.\\
Cate\_Tag: Education, Sports, Social.\\
Tool\_Tag: numbers, free\_nba, chuck\_norris.\\
API\_Tag: get\_random\_fact, get\_all\_stats, jokes\_search.\\
Solution\_Path:\\
Thought: get\_random\_fact\_for\_numbers, get\_all\_stats\_for\_free\_nba, get\_random\_fact\_for\_numbers, \\ jokes\_search\_for\_chuck\_norris, Finish.\\
\\

Now, Please make the tool using plan of below requests.\\
System: \{request\} \\
Solution\_Path:\\
         \hline
    \end{tabular}
    }
    \caption{The prompt for the solution path planning.}
    \label{prompt_5}
\end{table*}

\begin{table*}[!htb]
    \centering
    \resizebox{\textwidth}{!}{
    \begin{tabular}{|l|}
        \hline
         System: You are AutoGPT, you can use many tools(functions) to do the following task.\\
First I will give you the task description, and your task start.\\
At each step, you need to give your thought to analyze the status now and what to do next, with a \\ function call to actually excute your step. Your output should follow this format:\\
Thought:\\
Action:\\
Action Input:\\

\\
After the call, you will get the call result, and you are now in a new state.\\
Then you will analyze your status now, then decide what to do next...\\
After many (Thought-call) pairs, you finally perform the task, then you can give your finial answer.\\
Remember: \\
1.the state change is irreversible, you can't go back to one of the former state, if you want to restart the \\ task, say "I give up and restart".\\
2.All the thought is short, at most in 5 sentence.\\
3.You can do more then one trys, so if your plan is to continusly try some conditions, you can do one \\ of the conditions per try.\\
Let's Begin!\\
\\
Task description: You should use functions to help handle the real time user querys. Remember:\\
1.ALWAYS call "Finish" function at the end of the task. And the final answer should contain enough \\information to show to the user,If you can't handle the task, or you find that function calls always \\fail (the function is not valid now), use function Finish->give\_up\_and\_restart.\\
2.Do not use origin tool names, use only subfunctions' names.\\
You have access of the following tools: \{tool\_list\}\\
\\

Specifically, you have access to the following APIs: \{api\_list\}\\
User: \{Input\}\\
Assistant:
\\
         \hline
    \end{tabular}
    }
    \caption{The prompt for solution tree generation.}
    \label{prompt_6}
\end{table*}

\section{Details for ToolPlanner}\label{appendix_4}
\subsection{Stage1 SFT Model}\label{appendix_4_1}
\subsubsection{Prompt Design}
In this section, we show the details of the prompt design in ToolPlanner. 

The prompts of tag extraction, solution path planning and solution tree generation are shown in Table \ref{prompt_4}, \ref{prompt_5}, \ref{prompt_6}, respectively.

In the Stage 1 training phase, we use the tag list, solution path, and multi-round solution of 17,740 cases to finetune ToolPlanner. 

In the Stage 2 training phase, we use the 98,950 pairwise responses to further finetune ToolPlanner. 

In the test phase, ToolPlanner uses prompts for tag extraction and solution path planning to obtain the tag list and solution path. It then uses prompts for solution tree generation multiple times to generate the solution tree and final answer.

\begin{figure}[!htb]
  \centering
  \includegraphics[width=1\columnwidth]{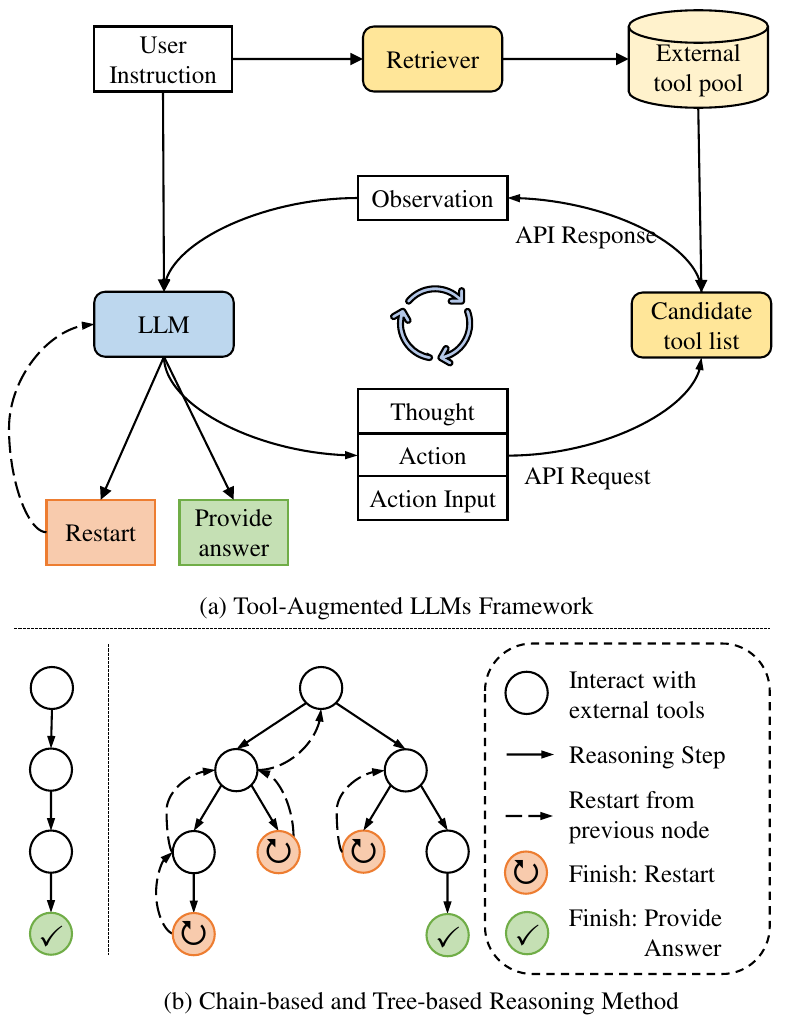}
  \caption{(a) The overview of a tool-augmented LLMs framework. (b) Two reasoning method with different structures. The tree-based reasoning method can generate a Restart node to terminate the current branch, and expand a previous node to continue the reasoning.}\label{figure3-1}

\end{figure}

\subsubsection{Inference Process of Tool-Augmented LLMs}\label{appendix_2}

In this section, we briefly introduced the framework for existing tool-augmented LLMs.
Existing tool-augmented LLM frameworks typically consist of an external tool pool, a retriever, and the main LLM.  The external tool pool contains all the tools and APIs that the framework can access. However, due to the limited context length of LLM, it is challenging to present all tool descriptions and usage examples to the model when there are numerous options available. Therefore, the retriever searches a limited subset of candidate tools and APIs related to the user instruction X from the external tool pool and provides them, along with their documentation, to LLM.

As shown in Figure \ref{figure3-1}, given the user instruction X and the candidate tool list, LLM needs to go through multiple rounds of reasoning and interaction with external tools to finally obtain a reasonable solution.  
In each round, LLM can perform the following operations: 1. Interact with external tools; 2. Thought; 3. Provide an answer; 4. Restart.


\begin{itemize}
\item \textbf{1. Interact with external tools:}
LLM can choose an external tool or API, generate and send an API request based on its documentation. The external tool can then process this request, generate a corresponding response, and send it back to LLM. The response from the external tool is considered an observation (O) of the LLM. Observations may be the correct information that the model needs or error logs generated during request processing. They are provided as history messages to the next round of the model. Such interactions can help the LLM expand its functionality and integrate external tools and services.
\item 
\textbf{2. Thought:} 
Based on user instructions and hisory messages, LLM can reason and describe its reasoning process, which is defined as the model's thought. Common thoughts include: 1) A task in the user instruction is still unfinished, and LLM needs to call a tool to complete this task. 2) There was an error in the previous response from an external tool, and LLM needs to regenerate its request sequence. 3) All tasks in the user instruction have been completed, and LLM can provide the results to the user.


\item 
\textbf{3. Provide an answer:} 
If the LLM believes that the task in the user instruction has been completed, it can use information from multi-round reasoning to provide an answer to the user. Furthermore, if the LLM believes that it cannot obtain more useful information from external tools, it can stop interacting with the external tools and describe the current reasoning process to the user.

\item \textbf{4. Restart (optional):} 
When an LLM determines that the current reasoning path cannot complete the user instruction, it can abandon the current path and restart from a previous round. 
Unlike early chain-based reasoning methods, many existing LLMs use tree or graph structures for their reasoning method. 
They treat each round of interaction as a node. 
When they determine that the current node is unlikely to generate a reasonable solution, they can return to a previous node and restart the reasoning process, even from the beginning. 
This method is more flexible because the model can switch between different branches to find the best solution.
Therefore, we use a tree-based reasoning method in ToolPlanner.
\end{itemize}

\begin{table*}[!htb]
    \centering
    \resizebox{\textwidth}{!}{
    \begin{tabular}{l|l}
         \hline
         \hline
Instruction & I'm planning a party for my friends and I want to include some fun jokes to entertain \\ 
& them. Can you provide me with ten jokes from the Jokes by API-Ninjas? Additionally, \\ & I would like to search for Chuck Norris jokes that mention cats. Can you find any for me? \\
\hline
Tag list  &  Cate\_Tag: Entertainment, Social.\\
& Tool\_Tag: jokes\_by\_api\_ninjas, chuck\_norris. \\
& API\_Tag: v1\_jokes, jokes\_random. \\
\hline
Solution Path & v1\_jokes\_for\_jokes\_by\_api\_ninjas, jokes\_random\_for\_chuck\_norris, Finish.\\
         \hline
         \hline
Solution Tree & \\
\hline
Action & v1\_jokes\_for\_jokes\_by\_api\_ninjas \\
Action Input & \{\}\\
\hline
Observation & \{"error": "", "response": "[ \{'joke': 'Why Does Snoop Dogg need an umbrella? \\
& For drizzle, my nizzle.' \} ] " \} \\
\hline
Action & jokes\_random\_for\_chuck\_norris \\
Action Input & \{"query": "cats"\}\\
\hline
Observation & \{"error": "", "response": "\{'total': 11, 'result': [\\ 
& \{'categories': [], 'created\_at': '2020-01-05 13:42:19.324003', \\ 
& 'icon\_url': 'https://assets.chucknorris.host/img/avatar/chuck-norris.png',\\ 
&  'id': 'g7OGUjc7T-WPMu5m\_-po\_Q', 'updated\_at': '2020-01-05 13:42:19.324003', \\ 
& 'url': 'https://api.chucknorris.io/jokes/g7OGUjc7T-WPMu5m\_-po\_Q', 'value': \\ 
& 'Chuck Norris decided he needed to have a pet. He looked at dogs and cats then ended \\ 
& up getting a pet alligator named Fluffy.'\}, \\ 
& \{'categories': [], 'created\_at': '2020-01-05 13:42:20.262289', \\ 
& 'icon\_url': 'https://assets.chucknorris.host/img/avatar/chuck-norris.png', \\ 
& 'id': 'QFBQ\_uDUTO2PTF3AfOhzuA', 'updated\_at': '2020-01-05 13:42:20.262289', \\ 
& 'url': 'https://api.chucknorris.io/jokes/QFBQ\_uDUTO2PTF3AfOhzuA', \\ 
& 'value': 'Creepers are scared of cats because Chuck Norris does not harm cats. Creepers \\ 
& know that this means that Chuck Norris likes cats and vice versa. Therefore, they know \\ 
& that if they harm a cat that Chuck Norris will harm THEM. They also know that Chuck \\ 
& Norris does not..."\} \\
         \hline
Answer & \{"return\_type": "give\_answer",\\ 
&  "final\_answer": "Sure! Here are ten jokes from the Jokes by API-Ninjas: \\ 
&  1. Why Does Snoop Dogg need an umbrella? For drizzle, my nizzle. \\ 
& 2. Chuck Norris decided he needed to have a pet. He looked at dogs and cats then ended \\ 
& up getting a pet alligator named Fluffy. \\ 
& Enjoy the party!"\} \\ 
         \hline
         \hline
    \end{tabular}
    }
    \caption{An example for the step-by-step inference process.}
    \label{prompt_10}
\end{table*}

\begin{table*}[!htb]
    \centering
    \resizebox{\textwidth}{!}{
    \begin{tabular}{|l|}
         \hline
         You have access of the following tools: \\ 1.jokes\_by\_api\_ninjas: Access a large collection of jokes from the internet. See more info at https://api-\\ninjas.com/api/jokes. \\
         2.chuck\_norris: chucknorris.io is a free JSON API for hand curated Chuck Norris facts. \\
         \hline
    \end{tabular}
    }
    \caption{An example for tool list.}
    \label{prompt_7}
\end{table*}

\begin{table*}[!htb]
    \centering
    \resizebox{\textwidth}{!}{
    \begin{tabular}{|l|}

         \hline
         API List: [\{\\
        \;\;\;\;"name": "v1\_jokes\_for\_jokes\_by\_api\_ninjas",\\
        \;\;\;\;"description": "This is the subfunction for tool "jokes\_by\_api\_ninjas", you can use this tool. The \\description of this function is: "API Ninjas Jokes API endpoint."",\\
        \;\;\;\;"parameters": \{\\
        \;\;\;\;\;\;\;\;"type": "object",\\
        \;\;\;\;\;\;\;\;"properties": \{ \},\\
        \;\;\;\;\;\;\;\;"required": [],\\
        \;\;\;\;\;\;\;\;"optional": [] \}\\
      \},\\
      \{\\
        \;\;\;\;"name": "jokes\_random\_for\_chuck\_norris",\\
        \;\;\;\;"description": "This is the subfunction for tool "chuck\_norris", you can use this tool.The description \\ of this function is: "Retrieve a random chuck joke in JSON format."",\\
        \;\;\;\;"parameters": \{\\
        \;\;\;\;\;\;\;\;"type": "object",\\
        \;\;\;\;\;\;\;\;"properties": \{ \},\\
        \;\;\;\;\;\;\;\;"required": [],\\
        \;\;\;\;\;\;\;\;"optional": [] \}\\
      \},\\
      \{\\
        \;\;\;\;"name": "Finish",\\
        \;\;\;\;"description": "If you believe that you have obtained a result that can answer the task, please call \\ this function to provide the final answer. Alternatively, if you recognize that you are unable to proceed \\ with the task in the current state, call this function to restart. Remember: you must ALWAYS call this \\ function at the end of your attempt, and the only part that will be shown to the user is the final answer, \\ so it should contain sufficient information.",\\
        \;\;\;\;"parameters": \{\\
        \;\;\;\;\;\;\;\;"type": "object",\\
        \;\;\;\;\;\;\;\;"properties": \{ "return\_type": \{  "type": "string", "enum": [ "give\_answer",  "give\_up\_and\_restart" \\ ] \}, "final\_answer": \{ "type": "string", "description": "The final answer you want to give the user. You \\should have this field if "return\_type"=="give\_answer""\} \},\\
        \;\;\;\;\;\;\;\;"required": [ "return\_type" ] \}\\
      \} ]\\
         \hline

    \end{tabular}
    }
    \caption{An example for API List.}
    \label{prompt_8}
\end{table*}



\subsubsection{Inference Process of ToolPlanner} \label{appendix_4_3}
In this section, we provide a step-by-step inference case to describe how ToolPlanner generates the tag list, solution path, solution tree, and final answer starting from a user instruction.

As shown in Table \ref{prompt_10},  a hybrid-level user instruction asks the model to provide some funny jokes, and specifies that these jokes should come from either API-Ninjas or Chuck Norris. 

First, ToolPlanner uses this instruction and the prompt in Table  \ref{prompt_4} for tag extraction. After tag extraction, ToolPlanner obtained tag lists at three different levels.

Then, ToolPlanner adds the tag lists to the model input and uses the prompt from Table \ref{prompt_5} to generate a solution path. After solution path planning, ToolPlanner obtained a three-step solution path: "v1\_jokes\_for\_jokes\_by\_api\_ninjas, jokes\_random\_for\_chuck\_norris, Finish".

From the tag lists, the model discovered that this task requires the Tools jokes\_by\_api\_ninjas and chuck\_norris, as well as the APIs v1\_jokes and jokes\_random from these two tools. The descriptions of these tools and APIs are shown in Table \ref{prompt_7} and Table \ref{prompt_8}, respectively.
The solution tree prompt from Table \ref{prompt_6} is completed with the instruction, tag lists, and solution path as model input, along with the above descriptions of the tools and APIs.

With the solution tree prompt, ToolPlanner interacts multiple times with the external tools to generate a solution tree and the final answer. Following the solution path, ToolPlanner first accesses the jokes\_by\_api\_ninjas tool and generates an API request. It observes that the tool returns a joke "Why Does Snoop Dogg need an umbrella? For drizzle, my nizzle." Then, the model accessed the chuck\_norris tool and set the query parameter in the API request to "cats". It observed that this tool returned two jokes. Finally, based on these observations, the model generated an answer and provided the jokes obtained from the two external tools to the user.

\newpage

\subsection{Stage 2 RL model}\label{appendix_RL}
In this section, we provide a detailed description of the process of extracting pairwise responses from the solution tree. As discussed in section \ref{section3.2}, the entire process consists of two steps: 1. Extracting solution paths and scoring them. 2. Extracting solution steps and pairing them up based on their reward scores.

\subsubsection{Reward}
\begin{figure}[!t]
  \centering
  \includegraphics[width=1\columnwidth]{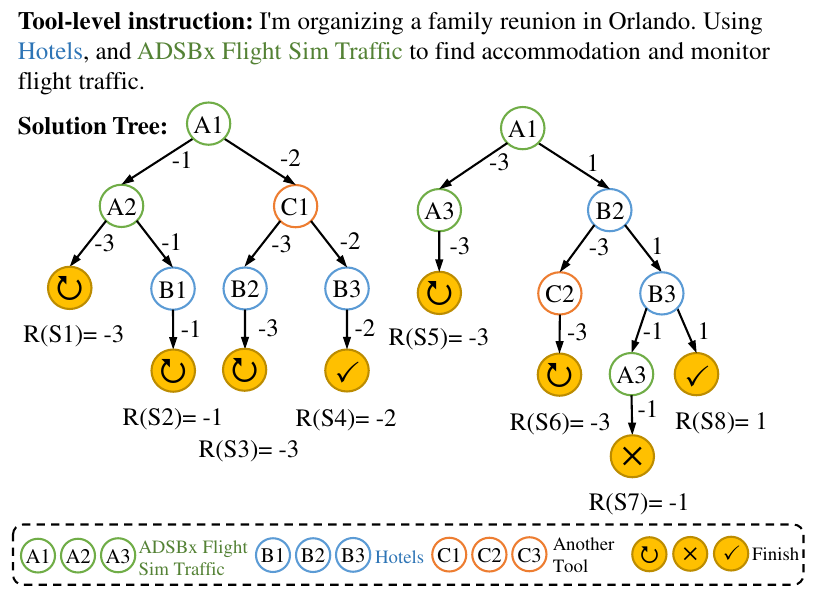}
  \caption{An example of a tool-level instruction and its two solution trees. }\label{figure4-3} 
\end{figure}

Figure \ref{figure4-3} shows an instruction and two corresponding solution trees. Each path from the root node to a leaf node is considered a solution. The figure contains a total of eight solution paths, namely S1, S2, $\ldots$, and S8.

We use task completion and instruction-following as metrics to score each solution. 

\begin{itemize}
\item \textbf{Task completion} measures whether the solution can successfully complete the task and finally provide a reasonable answer. Specifically, if the model finally decides to provide a response to the user, and this response is not meaningless, we mark the solution as "Pass" and set the pass reward to 1. Responses such as "Sorry, I couldn't find a suitable tool" are considered meaningless.  If the solution exceeds the maximum number of rounds or decides to restart, we mark it as "Not Pass". In a solution tree, at most one path may be marked as "Pass", namely, the rightmost one.

\item \textbf{Instruction-following} measures whether the solution follows the user's instructions. If the solution accesses and only accesses all categories, tools or APIs described in the instruction, mark it as "Match" and set match reward to 1, otherwise, mark it as "Not Match". If the level of the instruction is hybrid, we measure whether the solution matches the instruction at the API level. If the level of the instruction is tool, we measure whether the solution matches the instruction at Tool level. For example, if the user instruction explicitly mentions "assist me with tools from Mapping and Sports categories", we expect the solution to include tools from both categories and exclude tools from other categories. 
\end{itemize}

The reward score for each solution S can be calculated as follows:
\begin{equation}
\begin{split}
 {\rm R}(S)=\begin{cases}
 1 & \text{ if } S \in \text{Pass \& }S \in \text{Match} \\
 -1 & \text{ if } S \notin \text{Pass \& } S \in \text{Match} \\
 -2 & \text{ if } S \in \text{Pass \& } S \notin \text{Match} \\
 -3 & \text{ if } S \notin \text{Pass \& }  S \notin \text{Match}
\end{cases}
\end{split}
\end{equation}

In the two solution trees in Figure \ref{figure4-3}, S4 and S8 are marked as "Pass", while S2, S7, and S8 are marked as "Match". Therefore, only S8 has a reward of R(S8)=1.

For the i-th round of solution S, its reward score R(S\_i) is the highest reward score among all the solutions to which it belongs. 
Taking C1 inFigure \ref{figure4-3} as an example, C1 belongs to both solution S3 and solution S4. Since R(S4)=-2 and R(S3)=-3, R(C1|A1)=-2.

Since we use ranking loss to train the model, we only need to ensure that there is a difference in the ranking between different cases. Their score can be \{1,-1,-2,-3\} or \{3,2,1,0\}, both are acceptable. In this paper, for convenience, we choose \{1,-1,-2,-3\}.

Here, R(Not Pass \& Match) > R(Pass \& Not Match), it is because Match is more difficult than Pass in tool-using tasks.  After analyzing 4,443 seed training data from ToolBench, the distribution of these data on two metrics is shown in Table \ref{data_7}. The probability of "Not Match" is greater than "Not Pass".

\begin{table}[!tbh]
    \centering
    \resizebox{\columnwidth}{!}{
    \begin{tabular}{|c|c|c|c|}
         \hline
Pass \& &  Not Pass \& & Pass \& & Not Pass \&\\
Match &  Match  & Not Match & Not Match \\
         \hline
798 & 456 & 1288 & 1901  \\
         \hline
    \end{tabular}
    }
    \caption{Statistics on the distribution of reward metrics for 4,443 seed training data.}
    \label{data_7}
\end{table}

Given an instruction that requires the use of API1 and API2, "Not Pass \& Match" means that the model did not provide a valid answer due to length exceeding the limit or wrong judgment after correctly accessing API1 and API2, while "Pass \& Not Match" means that the model provides a valid answer without proper access to API1 and API2. If we punish "Not Match" less than "Not Pass", the model may tend to provide the answer immediately after successfully accessing the most related tool. In fact, this is also a common situation for "Pass \& Not Match" in the seed training data.

\subsubsection{Sampling and Ranking}
After annotating each node in the solution tree with a reward score, we can extract pairwise responses from it. When training ToolPlanner, each positive step is used to calculate the cross-entropy loss to fine-tune the model. Therefore, we only use nodes with a reward score of 1 as positive examples and extract nodes with the same history steps and negative reward scores as negative examples.

In Figure \ref{figure4-3}, only nodes belonging to the solution path S8 = "A1, B2, B3, $\checked$" are considered positive examples. For the node $\checked$, R($\checked$|A1,B2,B3) > R(A3|A1,B2,B3), therefore, ($\checked$, A3) is a pairwise response with (A1, B2, B3) as the history steps.
For the node B3, R(B3|A1,B2) > R(C2|A1,B2), therefore, (B3, C2) is a pairwise response with (A1, B2) as the history steps. For the node B2 with A1 as the history step, (B2, A2), (B2, C1), and (B2, A3) are three pairwise responses.

For nodes has no sibling nodes with a negative score, like A1, we sample and pair them with a negative example. There are three methods for sampling negative examples:
\begin{itemize}
    \item Select a Finish node to ensure that its path is marked as "Not Pass", such as $\circlearrowright$ and $\times$.
    \item Select a node from another tool to ensure that its path is marked as "Not Match", such as C1 and C2.
    \item If the history steps do not match the instruction, the $\checked$ node can be selected to end its path at "Pass \& Not Match".
\end{itemize}

\begin{table*}[!htb]
    \centering
    \resizebox{\textwidth}{!}{
    \begin{tabular}{|l|c|}
         \hline
         \hline
\textbf{Plausibility} & \textbf{Rate} \\
\hline
I'm planning a weekend getaway with my friends and I need some suggestions. Can you recommend & 3\\
 some vibrant cities with a lively nightlife in the United States? Also, provide me with a map of the & \\ 
selected cities and the nearest webcams for a glimpse of the atmosphere. & \\ 
\hline
I'm planning a weekend getaway with my friends and I need some suggestions. Using webcams & 2 \\
 \_travel, maptiles, and geocoder\_united\_states\_census\_bureau to help me find interesting locations  & \\ 
 and  activities. & \\ 
 \hline
I'm planning a weekend getaway with my friends and I need some suggestions. Using webcams& 1 \\
\_map\_ne\_lat\_ne\_lng\_sw\_lat\_sw\_lng\_zoom, getstandardmaptile, and geocoding\_and\_geolookup & \\ 
\_for\_an\_address  APIs, provide me with webcam locations, map tiles, and address details for & \\ 
potential destinations. &\\
\hline
\hline
\textbf{Conciseness} & \textbf{Rate} \\
\hline
I want to surprise my friend who is a cryptocurrency enthusiast with the latest market updates. Using & 3\\
currencyapi\_net and coinranking, provide me with current cryptocurrency information.& \\ 
\hline
I want to surprise my friend who is a cryptocurrency enthusiast with the latest market updates. Using & 2 \\
the APIs timeframe, convert, history, get\_coin\_markets, and get\_coin\_supply, provide me with & \\ 
recent cryptocurrency market information.& \\ 
\hline
I want to surprise my friend who is a cryptocurrency enthusiast with the latest market updates. Can & 1 \\
you provide me with the current prices and market information of the top 10 cryptocurrencies?& \\ 
Also, give me the historical rates between Bitcoin and Ethereum for the past week. Additionally, & \\ 
I would like to know the maximum supply and total supply of each coin.&\\
\hline
\hline
\textbf{Relevance} & \textbf{Rate} \\
\hline
I need to convert 1000 USD to EUR. Using currency\_exchange to find the conversion rate for me. & 3\\
\hline
I need to convert 1000 USD to EUR. Using exchange, getpercentage, jokes\_random APIs to provide & 2 \\
the conversion rate and a random joke.& \\ 
\hline
I need to convert 1000 USD to EUR. Can you also calculate the love percentage between John and & 1 \\
Alice? Lastly, could you share a random Chuck Norris joke? &\\
\hline
\hline
\textbf{Realness} & \textbf{Rate} \\
\hline
Provide the YEAR-END Top Artists - Female chart information for 2022. Using billboard\_api, & 3\\
deezer, and soundcloud to gather data and insights.& \\ 
\hline
Provide the YEAR-END Top Artists - Female chart information for the year 2022 on Billboard-API. & 2 \\
Using top\_artists\_female, radio, and song\_info APIs to get the chart data. &\\
\hline
Provide the YEAR-END Top Artists - Female chart information for the year 2022 on Billboard-API. & 1 \\
Fetch the radio details for the radio with the ID '123' on Deezer. Also, find the basic information& \\ 
of the song with the track URL 'https://soundcloud.com/user-977421934/the-phoenix' on Soundcloud. &\\
         \hline
         \hline
    \end{tabular}
    }
    \caption{Examples of instructions with different ratings.}
    \label{data_6}
\end{table*}

\section{Experiment}
\subsection{Main Metric}\label{appendix_metric1}
We used three metrics in main experiments: 

\textbf{Match Rate} measures the instruction following ability of LLM. If the solution accesses and only accesses all tags described in the user instructions, it is considered to match the instruction at the corresponding tag level.
Match Rate calculates the proportion of solutions that successfully match user instructions at a certain tag level.
When calculating the Match Rate of a fine-grained tag level, such as API, we also calculate the Match Rate of its parent tag levels, such as Tool and Category.
Taking an API-level instrucion as an example, if the solution generated by LLM only uses the tools mentioned in the instructions and uses all of these tools, then we consider this solution match the API-level instruction at the tool level. 
We evaluate the Category, Tool and API Match Rate for API-level and Hybrid-level instruction.

\textbf{Pass Rate} \cite{qin2023toolllm} measures whether the LLM is can successfully complete the task. If the LLM can successfully generate a “Finish” node with a reasonable answer within the maximum number of steps, we consider it pass the task.

\textbf{Win Rate} measures the quality of answers generated by LLM. We use ToolEval \cite{qin2023toolllm}  to compare the final answers generated by different LLMs and calculate the ratio at which ChatGPT prefers LLM answers over the golden answers from ToolBench. 

\subsection{Human Evaluation on Multi Granularity Instructions}\label{appendix_human}
To evaluate whether our multi granularity instruction mechanism can better reflect user behavior, we conducted a human evaluation using four metrics to compare user instructions at different levels.
\begin{itemize}
    \item \textbf{Plausibility:} This metric measures whether an instruction is fluent, complete, and makes sense in describing a user's intent. In other words, it measures whether the instruction conforms to the grammar and semantic rules of the language, and is like an executable task instruction.
    \item \textbf{Conciseness:} This metric measures whether an instruction is consistent and includes all necessary information. Are the instructions easy to understand and follow, or are they overly complicated and confusing?
    \item \textbf{Relevance:} This metric measures whether the "instruct" part of an instruction is clear and relevant to its statement. In other words, it determines whether the multiple tasks completed by different tools in the instructions are coherent and related to the statement sentences. For example, if the task statement is that the user needs to search for recipes, the command should not suddenly switch to calling "Playlist" from the music tool "Deezer".
    \item \textbf{Realness:} This metric measures whether an instruction aligns with the usage habits of real users, that is, whether the user is willing and able to use such instructions to instruct the model.
\end{itemize}

We randomly selected 100 instructions with different granularities and asked three crowdworkers to evaluate them. For each metric, we asked reviewers to rate the issues on a scale of 1-3 (with 3 being the best). Table \ref{data_6} provides examples of instructions with different ratings for each metric.

\begin{table*}[!htb]
  \centering
  \renewcommand{\arraystretch}{1.3}
  \caption{Human evaluation results on generated answers of different baselines.  }\label{table_human2}
  \resizebox{1.3\columnwidth}{!}
  {
    \begin{tabular}{cc|cccc}
    \hline
    Model A & Model B  & Both & A>B & B>A & Neither  \\
    \hline
    ToolPlanner  & ToolLlama-Tree  & 47.5\% & 26.5\% & 13.5\% & 12.5\% \\
	ToolPlanner & GPT-4 & 38\% &  41\% & 8.5\% & 12.5\%\\
    \hline
    \end{tabular}
  }
\end{table*}

Results of each human evaluation metric are presented in Table \ref{table_human1}.
We can see that:

\begin{itemize}
\item For plausibility, relevance, and realness, API-level instructions do not perform as well as others. Human evaluation has found that many API names are designed for developers and do not conform to natural language format or are irrelevant to the statement.
\item Hybrid-level instructions score low in conciseness due to their overly detailed and lengthy descriptions.

\item Tool-level instructions have achieved competitive performance compared to Hybrid-level instructions, with better relevance and realness but worse plausibility. This is because some tools have complex or oddly formatted names, which workers perceive as unnatural or not fluent for instructions that include them. 
When constructing Hybrid-level instructions, multiple APIs are first selected, and then statement and task instructions are constructed. This can result in some subtasks being irrelevant to their statements. Additionally, some Hybrid-level instructions may include specific API parameters, which can result in lower realness."

\item Category-level instructions have achieved the best or competitive performance in each metric. This is because they are very short, fluent, and easy for users to use. However, our dataset only contains 36 categories, which means it lacks diversity. Multiple solutions using different tools may correspond to similar category-level instructions. Therefore, we recommend using a combination of category-level, tool-level, and hybrid-level instructions.
\end{itemize}

\begin{table*}[!tb]
\centering
\resizebox{\textwidth}{!}{
\begin{tabular}{l|ccc|ccc|ccc|ccc}
 \hline \hline
\multirow{2}{*}{Model} & \multicolumn{3}{c|}{Category} & \multicolumn{3}{c|}{Tool} & \multicolumn{3}{c|}{API}  & \multicolumn{3}{c}{Hybrid} \\ \cline{2-13}
            & P      & R      & F1     & P      & R      & F1     & P     & R      & F1     & P      & R      & F1     \\ \hline
Retriever@1 & 0.89   & 0.4198 & 0.5705 & 0.99   & 0.4439 & 0.613  & 0.94  & 0.3333 & 0.4921 & \textbf{0.86}   & 0.305  & 0.4503 \\
Retriever@3 & 0.8413 & 0.75   & 0.793  & 0.9505 & 0.7758 & 0.8543 & 0.745 & 0.7872 & 0.7655 & 0.6734 & 0.7092 & 0.6908 \\
Retriever@5 & 0.7479 & 0.8538 & 0.7974 & 0.8472 & 0.87   & 0.8584 & 0.537 & 0.9255 & 0.6797 & 0.4949 & 0.8617 & 0.6287 \\ \hline
\textbf{Tag Extraction}         & \textbf{0.9906}   & \textbf{0.9953}  & \textbf{0.9929}  & \textbf{0.991}  & \textbf{0.9865} & \textbf{0.9888} & \textbf{0.9752} & \textbf{0.9752} & \textbf{0.9752} & 0.8486  & \textbf{0.8546}  & \textbf{0.8516} 
\\ \hline \hline
\end{tabular}
}
\caption{Compare the performance of Tag Extraction and Retriever in generating candidate lists. We show the performance of generating tags at the corresponding granularity for each instruction level.
}

    \label{result7}
\end{table*}

\subsection{Human Evaluation on Generated Answers}\label{appendix_human2}

As described in Section \ref{metric}, Win Rate uses ChatGPT to compare the generated answers of different baselines with the golden answers from ToolBench. To verify whether humans would make the same judgments as ChatGPT, we conducted human evaluations on the answers generated by different baselines.
We found two crowdsourcing workers who were provided with the final answers of two baselines on 100 Hybrid-level test cases, and asked them to compare and annotate whether the answers completed the instructions. The results of the human evaluation are presented in Table \ref{table_human2}. We can see that,

\begin{itemize}
\item Whether a solution passes or not has a significant impact on both the win rate and human evaluation. If the model does not generate a final answer or mentions in the final answer that it cannot use a certain tool, workers tend to annotate it as not having completed the instructions.
\item The performance of ToolPlanner and ToolLlama-Tree is basically consistent with the performance of the Win Rate metric. ToolPlanner performs better than ToolLlama-Tree.
\item GPT-4’s performance on human evaluation is worse than its performance on Win Rate. This is because sometimes even if GPT-4 does not successfully use the tool, it will provide a final answer and ask the user to provide more parameters. Model evaluation may consider such a response reasonable, but human evaluation may not.
\end{itemize}



\subsection{Comparison between Tag Extraction and Retriever}

We report the full version of the experiment comparing the Tag Extraction Mechanism and Retriever in Table \ref{result7}.

\end{document}